# Evaluation of Nuclear Microreactor Cost-competitiveness in Current Electricity Markets Considering Reactor Cost Uncertainties

Muhammad R. Abdussami*, Ikhwan Khaleb, Fei Gao, Aditi Verma*
University of Michigan, 2355 Bonisteel Blvd, Ann Arbor, MI 48109, USA

**Abstract**

This paper analyzes the cost-competitiveness of microreactors in contemporary electricity markets, emphasizing reactor cost uncertainties. Using a Genetic Algorithm (GA), we optimize key technical parameters of the microreactor life cycle, including reactor rated capacity, fuel enrichment, tail enrichment, refueling intervals, and discharge burnup, and determine the Levelized Cost of Energy (LCOE). The base case results are validated with Simulated Annealing (SA). By modeling variability in fuel cycle costs through Probability Distribution Functions (PDFs), we find optimal configurations for minimizing LCOE. Methodologically, the study introduces a novel optimization framework that integrates probabilistic cost modeling to evaluate microreactor LCOE. Empirically, it reveals how different cost uncertainties drive LCOE variability and influence perceptions of microreactor viability. Our findings show that despite cost uncertainties, microreactors can achieve competitive LCOE, ranging from $48.21/MWh to $78.32/MWh with Production Tax Credit (PTC). In general, higher reactor rated capacity, lower fuel enrichment, moderate tail enrichment, moderate refueling intervals, and higher discharge burnup are crucial for cost efficiency. The study demonstrates that overnight capital cost (OCC) significantly influences LCOE, while uncertainties in operational and maintenance (O&M) and fuel costs have a lesser impact. The results underscore the importance of design parameters, such as reactor capacity, fuel enrichment, tail enrichment, refueling duration, and discharge burnup, optimization before deploying microreactors to reduce LCOE. The results also highlight the economic viability of microreactors compared to traditional nuclear reactors, coal plants, and renewable energy technologies like offshore wind, biomass, and hydroelectric. The analysis reveals that energy policies like the PTC can reduce the LCOE by approximately 22-24%, even under microreactor cost uncertainties, highlighting its critical role in improving microreactor competitiveness. The study defines a realistic design space that highlights key trade-offs, supporting more targeted and transparent microreactor planning across diverse markets. The research provides policymakers, reactor designers, and energy planners valuable insights, promoting a sustainable and cost-effective energy future.

---

**Nomenclature**

| | |
|---|---|
| CF | Capacity factor |
| FTE | Full-time Equivalent Employee |
| GA | Genetic Algorithm |
| LCOE | Levelized Cost of Energy |
| LWR | Light Water Reactor |
| NRC | Nuclear Regulatory Commission |
| NG-fired CC | Natural Gas-fired Combined Cycle |
| OCC | Overnight Capital Cost |
| O&M | Operation and Maintenance |
| PDF | Probability Distribution Function |
| PTC | Production Tax Credit |
| SA | Simulated Annealing |
| SD | Standard Deviation |

**Variables and Parameters**

| | |
|---|---|
| $CF$ | Capacity factor (%) |
| $C_{cap}^{ann}$ | Annualized capital cost of microreactor |
| $C_{cap}^{tot}$ | Total capital cost of microreactor |
| $C_{dec}^{ann}$ | Annualized decommissioning cost of microreactor |
| $C_{dec}^{tot}$ | Total decommissioning cost of microreactor |
| $C_{FOM}^{ann}$ | Annualized fixed O&M cost of microreactor |
| $C_{VOM}^{unit}$ | Variable O&M cost of microreactor |
| $C_{YC}$ | Unit uranium cost (yellow cake) |
| $C_{UC}$ | Total uranium cost (yellow cake) |
| $C_{UCC}^{unit}$ | Unit uranium conversion cost |
| $C_{UCC}$ | Total uranium conversion cost |
| $C_{UE}^{unit}$ | Unit uranium enrichment cost |
| $C_{UF}^{unit}$ | Unit uranium fabrication cost |
| DB | Discharge burnup (MWd/kg) |
| $FTE$ | Number of FTE |
| $l$ | Uranium loss during conversion (%) |
| $LT_r$ | Lifetime of microreactor (Years) |
| $M_f$ | Number of kilograms per unit time of feed material |
| $M_p$ | Number of kilograms per unit time of product enriched |
| $M_t$ | Number of kilograms per unit time of uranium in the waste stream |
| $PTC$ | Production tax credit rate ($/MWh) |
| $P_{elec}$ | Rated capacity (electric) of microreactor ($MW_e$) |
| $r$ | Annual discount rate (%) |
| $S_{FTE}$ | Salary/compensation per FTE |
| $SP$ | Specific thermal power of the fuel (kW/kg) |
| $T_{refuel}$ | Refueling duration (Years) |
| $T_{PTC}$ | PTC duration (Years) |
| $T_H$ | Total number of hours in a year (8760) |
| $x_f$ | Weight fraction of U-235 in the feed material (%) |
| $x_p$ | Fuel enrichment (%) |
| $x_t$ | Tail enrichment (%) |
| $\eta$ | Thermal efficiency of the microreactor (%) |

## 1. Introduction

Nuclear power is widely recognized by experts as a key solution for sustainable energy, providing a reliable source of low-carbon electricity. Construction costs represent the largest share of total electricity generation costs for traditional large reactors, while the nuclear fuel cycle and Operation & Maintenance (O&M) costs contribute approximately 15-20% and 30-40%, respectively [1], [2]. However, its economic feasibility is a primary barrier for many countries in adopting nuclear power. The European Pressurized Reactor (EPR) at Flamanville began construction in 2007 with an initial budget of €3.3 billion and a planned completion in 2012. However, due to technical issues, regulatory challenges, and project management difficulties, the project experienced significant delays and cost overruns. By 2024, the reactor was connected to the grid, 12 years behind schedule, with costs escalating to approximately €13.2 billion, nearly four times the original estimate [3]. Similarly, Finland's Olkiluoto 3 EPR project faced substantial setbacks. Construction commenced in 2005 with an expected completion in 2009 and a budget of €3 billion. The project encountered numerous technical and legal challenges, leading to a 14-year delay. By its completion in 2023, the total investment cost had risen to around €5.8 billion, with some estimates suggesting the overall cost, including supplier expenses, reached up to €10 billion [4]. These examples demonstrate the persistent cost uncertainties associated with nuclear projects, ranging from capital investment to long-term operational expenses. Understanding and evaluating these uncertainties is essential for assessing the cost-competitiveness of emerging nuclear technologies like microreactors in current electricity markets.

Several studies have examined the potential of microreactors to alleviate energy poverty and enhance energy resilience, particularly in remote and underserved regions. While technical assessments suggest that microreactors could serve up to 70.9% of the unelectrified population, economic and governance limitations, such as regulatory challenges, security concerns, and institutional capacity, significantly constrain their practical deployment, reducing the viable market to less than 2% in many developing areas (e.g., Sub-Saharan Africa, parts of Asia, and some Latin American nations) [5]. Although microreactors could reach remote areas beyond grid infrastructure, their high levelized costs and the need for robust oversight institutions make them far less competitive than solar-battery systems in most regions [5], [6]. Despite their current high levelized costs ($140–$410/MWh), microreactors are viewed by some researchers as promising distributed energy resources for high-cost or isolated markets. Their ability to co-produce heat and hydrogen and complement renewable energy sources could enhance grid resilience, provided that licensing and market structures are reformed to support deployment [7]. However, other scholars remain skeptical. Critics argue that microreactors offer limited value beyond existing commercial technologies. Their safety, cost, and proliferation concerns outweigh potential benefits, especially when affordable renewable alternatives like solar and storage are readily available [8]. Still, some evaluations find that microreactors are cost-competitive with diesel generators and renewable sources in microgrids but not with large traditional nuclear plants [9]. This debate forms the basis of our investigation into the cost-competitiveness of microreactors compared to other currently available energy technologies.

Due to microreactors' developmental nature, studies estimating their costs remain limited. The projected costs vary significantly depending on assumptions related to policy, learning rates, and deployment contexts. Researchers broadly agree on the technical feasibility of microreactors but differ in assessing their economic competitiveness. Some studies focus on microreactor viability in specific use cases, especially off-grid applications [10], [11], [12]. Simulations using optimization tools such as HOMER have shown that microreactors could be competitive in remote microgrids, especially when considering emissions and reliability. However, these findings are tempered by the limitations of HOMER in accurately modeling microreactors. HOMER does not include a dedicated nuclear reactor model. Hence, users must simulate a microreactor by adapting an existing thermal generator model (such as natural gas or coal) and modifying only the economic parameters. However, this approach overlooks key technical characteristics of microreactors, which can significantly affect modeling accuracy [13], [14].

Comparative analyses between microreactors and conventional fossil or larger nuclear technologies reveal promising economic scenarios under specific conditions. Nuclear batteries, a class of microreactors, have been shown to be potentially cost-competitive with natural gas generators, especially when carbon pricing mechanisms are considered. Projected electricity costs for these systems range between \$70–115/MWh, with co-generated heat priced at \$20–50/MWh, suggesting viability in decarbonized energy markets [15]. Similarly, very small modular reactors (vSMRs) or microreactor, such as the Westinghouse eVinci, demonstrate strong economic performance in targeted applications. In a zero-emission city case study, the eVinci design can achieve a notably low levelized cost of electricity at \$24.6/MWh, outperforming other small modular reactor designs like SMART and IMSR. These findings indicate that vSMRs or microreactors, particularly when integrated with renewable systems, may offer a cost-effective solution for energy supply in remote or infrastructure-limited environments [16].

The cost-competitiveness of microreactors is highly sensitive to geographic and contextual factors. Analyses across diverse scenarios, ranging from off-grid to grid-connected applications, demonstrate that infrastructure availability, market conditions, and regional deployment contexts significantly influence economic viability. Certain markets, including isolated operations, distributed energy systems, resilient urban infrastructure, disaster relief, and marine propulsion, have been identified as well-suited for microreactor deployment due to their unique operational needs and constraints [17], [18], [19], [20], [21].

While several studies incorporate basic sensitivity analyses, most do not account for the full range of cost variability. A key gap in literature is the limited use of probabilistic methods to model uncertainty. For example, uranium pricing alone can significantly impact overall cost estimates, highlighting the importance of incorporating stochastic modeling techniques in fuel cycle assessments. Probabilistic approaches offer a more realistic evaluation of economic risk and are essential for informing policy and investment decisions under uncertainty [22]. Overall, while existing research provides valuable insights into the economics of microreactors, gaps remain in accurately modeling the full-spectrum of cost uncertainties and comparing microreactors against other technologies under realistic, variable conditions.

This study has two-fold contributions. Methodologically, the paper presents a novel framework to evaluate the cost-competitiveness of nuclear microreactors by modeling and optimizing their Levelized Cost of Energy (LCOE) while explicitly accounting for cost uncertainties across the fuel cycle. Unlike prior studies that often rely on fixed or aggregated cost assumptions, this study formulates detailed Probability Distribution Functions (PDFs) for key cost components, including capital, O&M, and fuel cost. The research incorporates them into a constrained optimization model solved using Genetic Algorithms (GA) and validated with Simulated Annealing (SA). A significant methodological advancement is the distinction between design variables that must be fixed prior to mass production (e.g., refueling schedule, reactor capacity, fuel enrichment, burnup, etc.) and those that can be fine-tuned based on deployment context (e.g., capital costs, O&M cost, and fuel cost). The study provides a realistic “design space” that reflects the trade-offs available to designers and policymakers across different markets, enabling more targeted and transparent microreactor planning.

Empirically, this analysis offers new insights into how different components of reactor cost uncertainty contribute to overall variation in microreactor LCOE. By reconstructing comparative LCOE charts under uncertainty, this study clarifies why debates persist about microreactor viability. Rather than settling the debate, the study shows that such disagreements are rooted in the non-trivial variability of cost assumptions. The findings emphasize the need for greater transparency and realism in microreactor cost estimation to avoid repeating the overly optimistic expectations that have plagued large reactor projects.

The paper organization is as follows: Section 2 describes the research methodology, and Section 3 illustrates the optimization problem formulation, including the objective function, constraints, boundaries, and decision variables. Section 4 explains the results, and Section 5 provides the conclusion.

## 2. Research Methodology

The research methodology is depicted in Fig. 1. Initially, comprehensive data, such as different reactor costs and PDFs, have been gathered to establish a cost model for the life cycle of a microreactor. We assume a generic microreactor design representative of uranium-fueled, thermal-spectrum systems with TRISO or similar fuel types. The model does not reflect a specific vendor or design but captures typical technical and economic characteristics of current microreactor concepts. While the assumptions provide a broad baseline, results may vary for microreactors with alternative fuel types (e.g., metallic fuel), cooling technologies (e.g., gas, molten salt), or deployment models (e.g., mobile vs. stationary), and should be interpreted accordingly.

Table 1 details the microreactor's cost parameters and their respective values. The Overnight Capital Cost (OCC) is treated as an "nth-of-a-kind" product and encompasses the expenses related to the fabrication, transportation, installation, and connection of the reactor and power conversion unit, in addition to site preparation and the construction of service buildings. Fixed Operation and Maintenance (O&M) costs cover regulatory fees from the Nuclear Regulatory Commission (NRC), inspections, insurance premiums, and property taxes. The total compensation for Full-time Equivalent Employee (FTE) staff includes benefits and taxes. The cost of uranium is based on a conservative estimate of the price of yellowcake. It is assumed that the cost of fuel fabrication is double that of traditional Light Water Reactor (LWR) fuel fabrication [15]. The cost for spent fuel disposal is estimated to align with the current fee for the disposal of spent nuclear fuel in the U.S. Decommissioning costs are incurred at the end of the project. The Capacity Factor (CF) assumes approximately four months of operational downtime per five years, which accounts for the unexpected loss of availability [15]. Additionally, incentives from the Production Tax Credit (PTC), as outlined in the Inflation Reduction Act (IRA), are considered in the model.

Then, a Probability Distribution Function (PDF) is formulated for each identified cost parameter of the microreactor. Table 2 lists these cost parameters along with their corresponding PDFs. The cost parameters vary widely due to factors such as technology, geography, policy, and energy markets. For example, spent fuel cost is well-defined in the U.S.; therefore, it is kept fixed throughout the study, with no uncertainty considered. The PDFs for uranium cost, uranium conversion cost, uranium enrichment cost, and fuel fabrication cost are assumed based on the findings in [22]. The remaining PDFs for the cost parameters are assumed to follow a "Uniform" distribution as a conservative assumption due to the lack of literature depicting PDFs for these costs.

In the next step, "Roulette Wheel" selection method chooses a single cost parameter of the optimization problem. This method models a selection as a roulette wheel where each segment size is proportional to its fitness. The algorithm involves potentially normalizing the fitness values to ensure all are non-negative and scaling them if any value is less than one. A cumulative probability distribution is then constructed, where an individual's probability of being selected is cumulative and proportional to its fitness. A random draw determines the selection by matching a randomly generated number with the intervals in the cumulative distribution. This method ensures that individuals with higher fitness have a higher probability of being selected, thus mimicking the natural selection by favoring the propagation of advantageous traits [23].

Then, an optimization problem is formulated to determine the LCOE for a microreactor. LCOE is a metric used to assess the average total cost to build and operate a power-generating asset over its lifespan, divided by the total energy output over that lifespan. The LCOE calculation considers various factors, including the initial capital costs (such as construction and commissioning), ongoing operation and maintenance costs, the cost of fuel, the expected annual energy production, the project's lifespan, incentives (e.g., PTC), and the discount rate (which reflects the time value of money). Depending on the energy technology being assessed, it can also factor in the costs related to environmental impacts (if any), decommissioning, and waste disposal. The details of the optimization problem including objective function, decision variables, constraints, and boundaries are presented in Section 3. The optimization problem is solved using a nature-based optimization algorithm called Genetic Algorithm (GA) in MATLAB environment, and the results obtained in the GA are validated using another optimization algorithm called Simulated Annealing (SA) in MATLAB. GA is well-suited for solving complex, nonlinear, multi-variable optimization problems with constraints. It can efficiently search for a large design space and avoid getting trapped in local minima, making it ideal for optimizing microreactor design variables. GA starts with a randomly generated population of potential solutions, each encoded as a chromosome. These individuals are evaluated using a fitness function, with higher scores indicating better solutions. Selection is based on fitness, favoring superior individuals for reproduction through roulette wheel selection. During crossover, genetic material is exchanged between pairs of individuals to produce varied offspring, while mutation introduces random changes to genes, maintaining diversity within the population. The process iterates, replacing older generations with newer ones, until a termination condition, such as a maximum number of generations or an acceptable fitness level, is met [24]. Since GA is a nature-based optimization algorithm that does not guarantee a global minimum, we run each of the scenarios twenty times and extract the lowest LCOE and their corresponding decision variable values to avoid cases where solutions get stuck in local minima. We chose twenty runs to balance computational cost and solution reliability. In preliminary tests, we observe that beyond twenty runs, the improvement in LCOE becomes marginal, indicating diminishing returns. Thus, twenty iterations provide consistent results while keeping the computation time reasonable. SA is a probabilistic optimization technique inspired by the annealing process in metallurgy, where a material is heated and then slowly cooled to remove defects and optimize its structure. SA operates by iteratively exploring the solution space, probabilistically accepting or rejecting new solutions based on a temperature parameter that gradually decreases over time [25]. SA can also escape local minima by accepting worse solutions early in the process. Using SA ensures robustness and confirms that GA has not converged prematurely on suboptimal solutions.

Finally, the objective function value and the corresponding decision variables are recorded for each scenario. A total of 100 scenarios are formulated using the "Roulette Wheel" selection method to select values for the cost parameters randomly. We select 100 scenarios to ensure a statistically meaningful exploration of the cost uncertainty space while maintaining computational efficiency. This number provides a diverse range of outcomes, allowing us to observe variability and trends in LCOE without excessive simulation time. Initially, the optimization process is run using the nominal values of the cost parameters without considering any uncertainty. In the subsequent analysis, cost uncertainty is introduced into the optimization process to evaluate its impact compared to the base case.

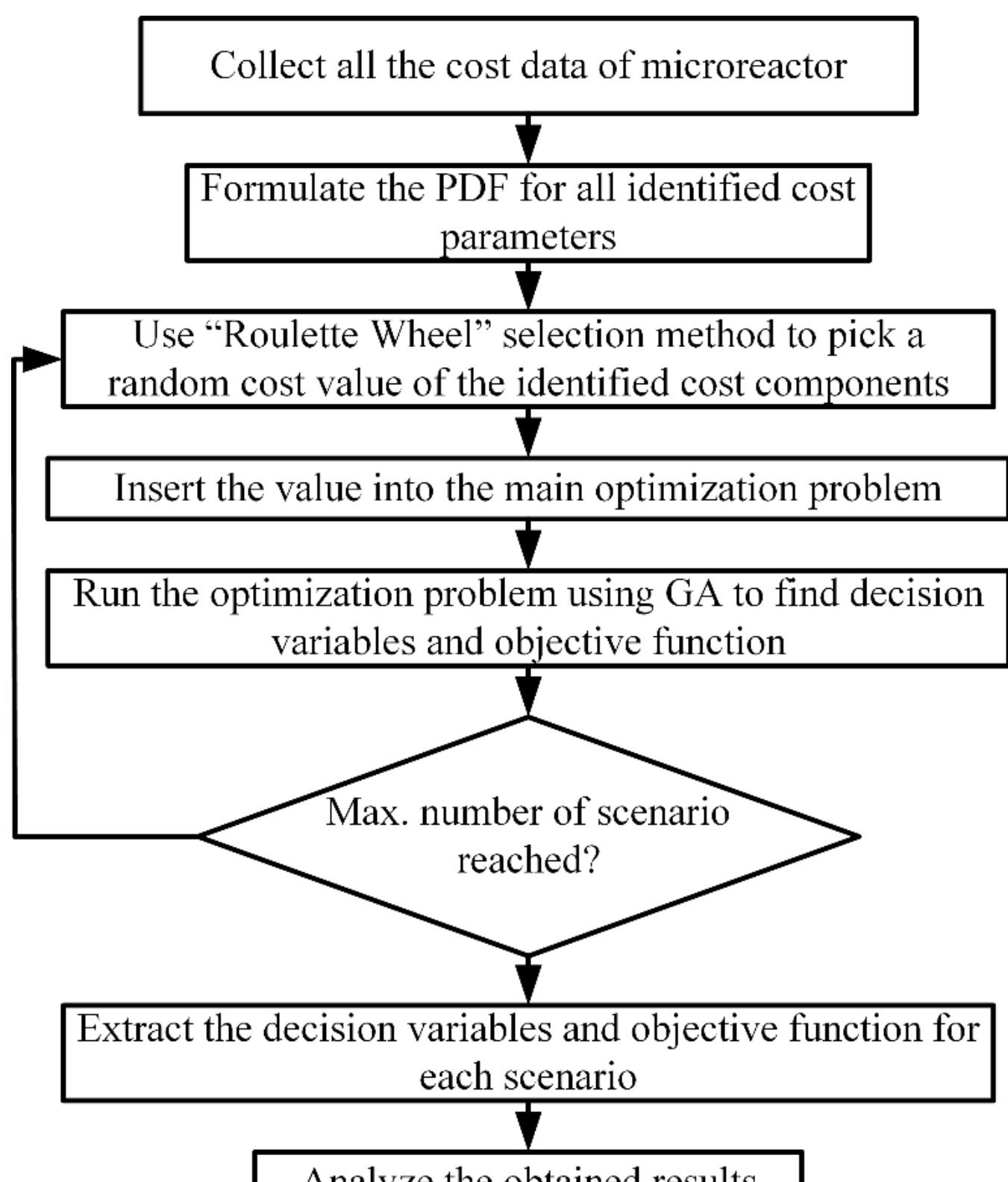

Fig. 1: Flowchart of the research methodology

Table 1: Techno-economic parameters of a generic microreactor technology [15], [26]

| **Parameters** | **Value (base case)** |
|---|---|
| OCC | $3,000\ \$/KW_e$ |
| Fixed O&M cost | 0.5 M $/year |
| Variable O&M cost | 2.07 $/MWh |
| No. of FTE for O&M | 5 |
| Compensation per FTE | 150,000 $/year |
| Uranium cost | 40 $/lb of $U_3O_8$ |
| Uranium conversion cost | $6\ \$/kg_U$ |
| Uranium enrichment cost | 160 $/SWU |
| Fuel fabrication cost | $500\ \$/kg_U$ |
| Spent fuel disposal cost | 1 M$/kWh |
| Decommissioning cost | $7500\ \$/KW_e$ |
| Capacity factor (CF) | 93% |
| Efficiency | 35% |
| Refueling and servicing downtime | 6 months |
| PTC | 25$/MWh (for first 10 years) |
| Reactor lifetime | 20 years |
| Real discount rate | 5% |
| Inflation rate | 2% |
| Project lifetime | 20 years |

Table 2: PDF and corresponding values of fuel cycle cost

| **Parameters** | **PDF** [22] | **Max** | **Min** | **Nominal** |
|---|---|---|---|---|
| Capital cost ($\$/KW_e$) | Uniform | 4000 [27] | 2500 | 3000 |

| No. of FTE for O&M | Uniform | 10 | 3 | 5 |
|---|---|---|---|---|
| Compensation per FTE ($/year) | Uniform | 225,000 | 120,000 | 150,000 |
| Fixed O&M cost ($/year) | Uniform | 0.75M | 0.40M | 0.5 M |
| Variable O&M cost ($/MWh) | Uniform | 2.5 | 2.0 | 2.07 |
| Uranium cost ($/lb) | Triangular | 156 | 84 | 104 |
| Uranium conversion cost ($/kg$_U$) | Uniform | 10 | 4 | 6 |
| Uranium enrichment cost ($/SWU) | Uniform | 240 | 125 | 160 |
| Fuel fabrication cost ($/kg$_U$) | Triangular | 750 | 400 | 500 |

## 3. Problem Formulation

### 3.1. Objective function

The objective function of this study is to minimize the LCOE, presented by eq. (1).

$$LCOE = \frac{Total\ discounted\ annual\ expenses}{Total\ annual\ energy\ production} \tag{1}$$

The LCOE incorporates all cost types listed in Table 1. The numerator of eq. (1) is the sum of all types of annualized cost. The annualized capital cost of the microreactor is calculated using eq. (2).

$$C_{cap}^{ann} = C_{cap}^{tot}\left\{\frac{r(1+r)^{LT_r}}{(1+r)^{LT_r}-1}\right\} \tag{2}$$

The fixed O&M cost is already expressed in the form of annualized cost. Therefore, total annualized O&M costs can be given by eq. (3).

$$C_{O\&M}^{ann} = (FTE \times S_{FTE}) + C_{FOM}^{ann} + \left(C_{VOM}^{unit} \times CF \times T_H \times P_{elec}\right) \tag{3}$$

Total annualized fuel cost is the sum of uranium cost, uranium conversion cost, uranium enrichment cost, and uranium fabrication cost. Total uranium cost can be expressed as follows by eq. (4-7).

$$C_{UC} = \frac{C_{YC} \times M_f}{1-l} \tag{4}$$

$$M_f = \frac{x_p - x_t}{x_f - x_t} M_p \tag{5}$$

$$M_p = \frac{P_{elec}}{SP \times \eta} \tag{6}$$

$$SP = \frac{DB}{T_{refuel} \times CF \times 365} \tag{7}$$

Total uranium conversion cost can be written as follows using eq. (8).

$$C_{UCC} = C_{UCC}^{unit} \times M_f \tag{8}$$

Total enrichment cost of uranium can be expressed as follows by eq. (9-14) [28]. Eq. (5) is the feed rate, while eq. (14) is the waste rate.

$$C_{UE} = M_p \times SWU \times C_{UE}^{unit} \tag{9}$$

$$SWU = \frac{M_p V_p + M_t V_t - M_f V_f}{M_p} \tag{10}$$

$$V_p = \left(2x_p - 1\right)\ln\frac{x_p}{1-x_p} \tag{11}$$

$$V_t = (2x_t - 1)\ln\frac{x_t}{1-x_t} \tag{12}$$

$$V_f = \left(2x_f - 1\right)\ln\frac{x_f}{1-x_f} \tag{13}$$

$$M_t = \frac{x_p - x_f}{x_p - x_t} M_f \tag{14}$$

Total uranium fabrication cost can be determined as follows using eq. (15).

$$C_{UF} = M_p \times C_{UF}^{unit} \tag{15}$$

Therefore, the annualized fuel cost of a microreactor can be written as follows by eq. (16).

$$C_{fuel}^{ann} = (C_{UCC} + C_{UE} + C_{UF})\left\{\frac{r(1+r)^{T_{refuel}}}{(1+r)^{T_{refuel}}-1}\right\} \tag{16}$$

The annualized decommissioning cost of microreactor can be calculated using eq. (17).

$$C_{dec}^{ann} = C_{dec}^{tot}\left\{\frac{r}{(1+r)^{LT_r}-1}\right\} \tag{17}$$

The annualized incentives obtained from PTC for microreactors can be written as follows using eq. (18).

$$C_{PTC}^{ann} = PTC \times CF \times T_H \times P_{elec} \times \frac{(1+r)^{T_{PTC}}-1}{r(1+r)^{T_{PTC}}} \times \left\{\frac{r(1+r)^{LT_r}}{(1+r)^{LT_r}-1}\right\} \tag{18}$$

### 3.2. Decision variables

Although many parameters influence the microreactor LCOE, many of these values remain fixed regardless of technology and design changes. For example, parameters such as O&M compensation and decommissioning cost are treated as fixed because they are influenced more by external market conditions or regulatory frameworks than by design choices. Therefore, we only consider the variable design parameters as decision variables that designers can actively optimize. The decision variables for this study are reactor rated capacity ($MW_e$), fuel enrichment (%), tail enrichment (%), refueling duration (years), and discharge burnup (MWd/kg), as listed below.

$$x = \left[P_{elec}, x_p, x_t, T_{refuel}, \text{DB}\right] \tag{19}$$

### 3.3. Constraints

The optimization is subject to the following nonlinear equality constraint which ensures the mass balance in the uranium fuel cycle [28]. It relates the technical and economic parameters of fuel cycle analysis, as presented by eq. (20).

$$x_f M_f = x_p M_p + x_t M_t \tag{20}$$

Additionally, fuel enrichment, burnup, and refueling duration are critical parameters that directly influence microreactors' fuel utilization efficiency and operating cycle. These parameters significantly impact the LCOE by affecting fuel cost, refueling frequency, and spent fuel management. Therefore, another empirical equality constraint is employed in this study, as presented by eq. (21) [29]. Eq. (21) is derived for PWR/VVER-type thermal reactors. Since it is challenging to use reactor-specific constraints relating to fuel enrichment, burnup, and refueling time, we assume that this simplified empirical relationship is a reasonable approximation for a typical microreactor operating in the thermal spectrum and fueled by uranium, including designs that utilize TRISO.

$$DB = 14.8x_p - \frac{SP \times 365 \times T_{refuel}}{1000} \tag{21}$$

### 3.4. Boundaries

The decision variables are constrained within specified bounds. The upper and lower limits set in the optimization problem are presented below [15], [28], [30], [31].

$$1 \le P_{elec} \le 20 \qquad x_p \epsilon R^+ \tag{22}$$

$$5 \le x_p \le 20 \qquad x_p \epsilon R^+ \tag{23}$$

$$0.2 \le x_t \le 0.3 \qquad x_t \epsilon R^+ \tag{24}$$

$$2 \le T_{refuel} \le 10 \qquad T_{refuel} \epsilon R^+ \tag{25}$$

$$15 \le \text{DB} \le 30 \qquad \text{DB} \epsilon R^+ \tag{26}$$

## 4. Results

Using the nominal values of the cost parameters, the optimization provides the microreactor LCOE as \$51.79/MWh. Fig. 2 compares the microreactor LCOE with other currently available energy technologies, including the PTC if applicable [32]. The figure shows that microreactors with PTC have a lower LCOE than offshore wind, biomass, Ultra-Supercritical (USC) coal, Traditional Advanced (TA) reactors, and hydroelectric. However, microreactor LCOE with PTC is higher than hybrid solar, onshore wind, Natural Gas (NG)-fired Combined Cycle (CC), geothermal, and standalone solar.

Table 3 presents the optimal values of the objective function and decision variables as determined by GA for base case analysis (without consideration of uncertainty). GA suggests a higher rated electric capacity of microreactors to lower the LCOE. It happens due to the complex trade-off between the cost and efficiency of microreactors, which can be comprehended by the equations presented in section 3.1. GA identifies 5% fuel enrichment as the most cost-effective. Typically, higher fuel enrichment increases costs by raising the Separative Work Units (SWU). Conversely, a higher tail enrichment percentage lowers the LCOE, leading the GA to select a value of 0.2913%. GA selects an optimal refueling period of around 6.24 years, balancing cost implications and operational practicalities. While longer refueling durations can reduce LCOE, they also lead to higher interest over time and increased required fuel mass, which raises costs. Therefore, the GA selects a mid-range refueling duration. Lastly, the GA favors higher discharge burnup, which enhances the fuel's specific power, reducing the needed fuel mass and thus lowering enrichment and fabrication costs.

The base case result is validated using SA, which is also presented in Table 3. The results from GA and SA are nearly identical, with minor discrepancies due to both being metaheuristic optimization algorithms. The results presented in the later sections are obtained only from GA.

Fig. 3 illustrates the cost breakdown that contributes to the LCOE for microreactors. It shows that the OCC contributes the highest to the LCOE at 60.8%, similar to traditional nuclear reactors. Fuel costs account for about 23.2% of the total LCOE, comparatively higher than conventional reactors, where fuel costs range between 15-20%. Conversely, microreactors' O&M costs represent around 16% of the LCOE, lower than the 30-40% typically observed in traditional reactors. The study also reveals the significant impact of the PTC on nuclear technologies. The results indicate that the PTC can reduce the overall LCOE by roughly 23%. Therefore, supportive nuclear energy policies are essential for enabling new nuclear technologies to compete effectively in electricity markets. According to [33], the LCOE of microreactors can range from $80/MWh to $340/MWh (2019 USD) for NOAK designs, depending on factors such as rated power capacity, the number of units deployed, and specific cost assumptions. This wide range highlights the importance of optimizing design parameters, as doing so can significantly reduce costs and enhance the economic viability of microreactors in competitive energy markets.

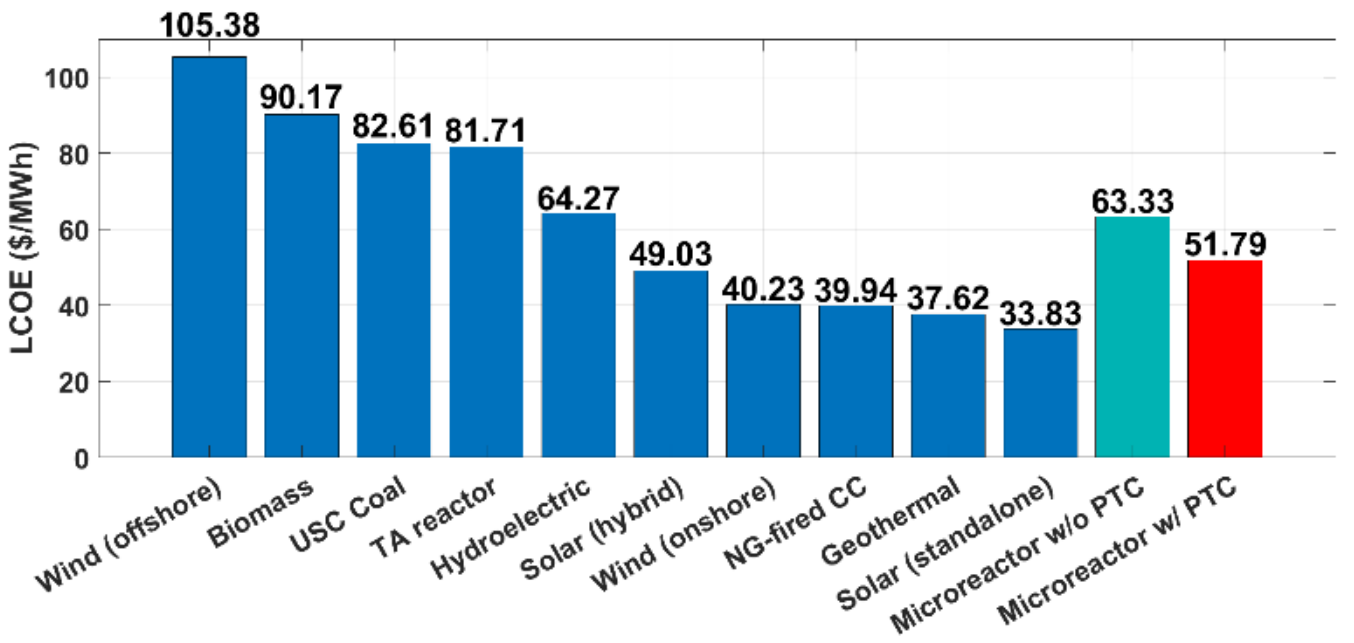

Fig. 2: LCOE comparison of different energy technologies (Base case without cost uncertainty)

Table 3: Optimal values of the decision variables (base case)

| Parameters | Value (GA) | Value (SA) |
|---|---|---|
| LCOE ($/MWh) | 51.79 | 52.01 |
| Reactor capacity ($MW_e$) | 19.13 | 19.2 |
| Fuel enrichment (%) | 5 | 5 |
| Tail enrichment (%) | 0.2913 | 0.2993 |
| Refueling duration (Years) | 6.24 | 6.72 |
| Discharge burnup (MWd/$Kg$) | 30 | 30 |

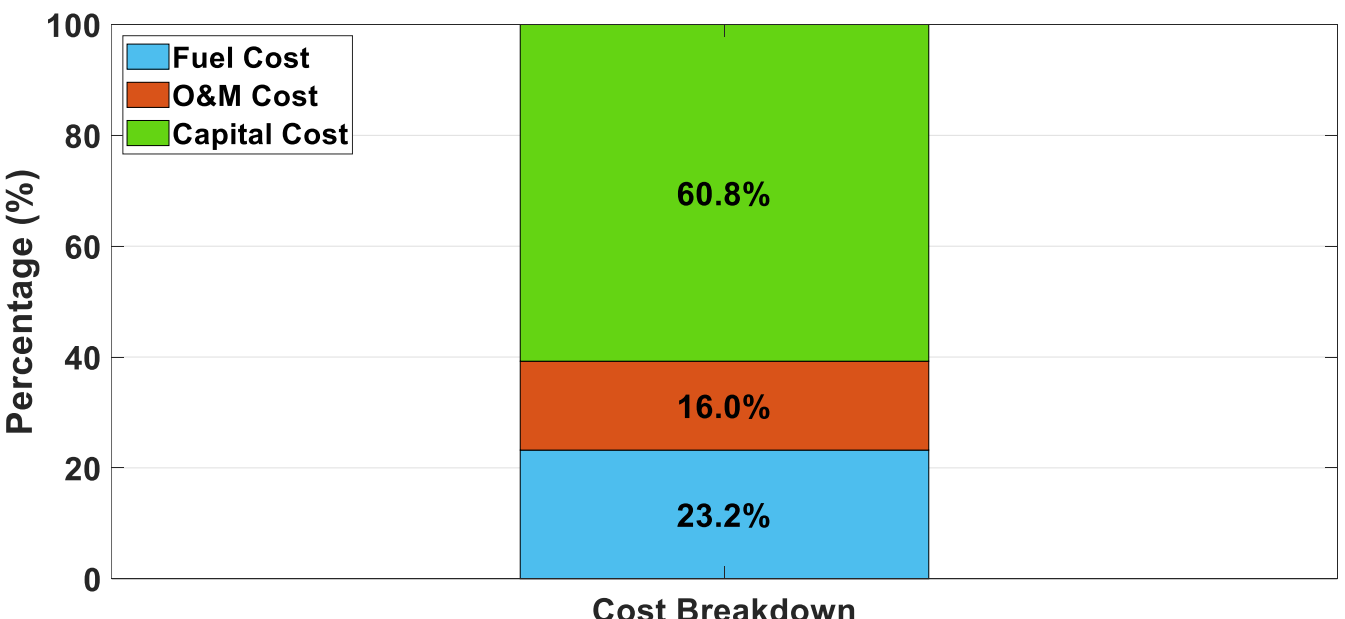

Fig. 3: LCOE contribution by different types of cost

In the following section, we assess the impact of cost uncertainty on LCOE with PTC and design parameters (e.g., reactor capacity, fuel enrichment, tail enrichment, refueling duration, and discharge burnup) in four different ways: the impact of overall cost uncertainty on LCOE and design parameters, the impact of OCC uncertainty on LCOE and design parameters, the impact of O&M cost uncertainty on LCOE and design parameters, and the fuel cost uncertainty on LCOE and design parameters, as illustrated in Fig (4–13). First, we investigate the impact of overall cost uncertainty of the different cost parameters listed in Table 2. The optimization results after introducing all types of cost uncertainty are shown in Figs. 4 and 8. The results indicate that LCOE can vary from $48.21/MWh to $78.32/MWh, with a Standard Deviation (SD) of 7.2. This significant SD suggests that LCOE is highly sensitive to cost uncertainty. It is challenging to conclude definitively whether microreactors are more cost-effective than other energy sources mentioned in the base case optimization. However, the LCOE of microreactors is consistently lower than that of offshore wind, biomass, USC coal, and TA reactors, regardless of cost uncertainty, as depicted in Fig. 4. Additionally, under certain reactor cost parameters, the LCOE of microreactors could be lower than that of hydroelectric and hybrid solar, as illustrated in Fig. 4. The findings further suggest that, under comprehensive uncertainty analysis, the PTC can lower the LCOE by between 22.76% and 24.37%.

The first quartile (Q1), median (Q2), and third quartile (Q3) values of LCOE in Fig. 4 offer deeper insights into the impact of cost uncertainty on microreactor LCOE. We obtain a Q1 value of $55.38/MWh, which signifies that 25% of the scenarios yield an LCOE below this value. A median value of $61.13/MWh indicates that 50% of the scenarios have an LCOE below this value. Similarly, a Q3 value of $66.43/MWh demonstrates that 75% of the scenarios result in an LCOE below this value, with the remaining 25% above it. These quartile values underscore that in 75% of scenarios, the LCOE is roughly more cost-effective than offshore wind, biomass, USC coal, and TA reactors, thereby highlighting the potential cost-effectiveness of microreactors. The detailed statistical results of the optimization problem and the design parameter are presented in Table A1 (Appendix: A) and Figs. A1-A5 (Appendix: B). Reactor capacity varies between 19.91 $MW_e$ to 12.23 $MW_e$ with an SD value of 1.47, as shown in Fig. 9. Fuel enrichment and discharge burnup remain constant at 5% and 30 MWd/kg, respectively, depicted in Figs. 10 and 13, indicating no variation in these parameters. Tail enrichment varies widely within the maximum and minimum limits, presented in Fig. 11. Refueling duration ranges from 9.95 to 2.48 years, with an SD of 1.96, shown in Fig. 12.

The cost parameters corresponding to the maximum LCOE of $78.32/MWh include a capital cost of $3969.7/$\mathrm{kW_e}$, uranium cost $118.18/lb, uranium conversion cost of $9.7/$\mathrm{kg_U}$, uranium enrichment cost of $190.05/SWU, and fuel fabrication cost of $587.37/$\mathrm{kg_U}$. Additionally, this scenario involves nine FTE for O&M with a compensation of $187,878.80/year, a fixed O&M cost of $0.746M/year, and a variable O&M cost of $2.08/MWh. In contrast, the parameters for the minimum LCOE of $48.21/MWh are noticeably different. These include a lower capital cost of $2621.21/$\mathrm{kW_e}$, uranium cost $110.18/lb, uranium conversion cost of $4.78/$\mathrm{kg_U}$, enrichment cost of $149.4/SWU, and fuel fabrication cost of $608.59/$\mathrm{kg_U}$. The scenario assumes six FTEs, each earning $182,575.80/year, along with a fixed O&M cost of $0.467M/year, and a variable O&M cost of $2.35/MWh.

At first glance, these comparisons suggest that minimizing OCC consistently leads to a lower LCOE. However, this is not always the case. For instance, in another scenario with a capital cost of $2560.6/$\mathrm{kW_e}$ (very close to the minimum value of OCC from Table 2), uranium cost $107.27/lb, uranium conversion cost of $5.27/$\mathrm{kg_U}$, enrichment cost of $191.2/SWU, and fuel fabrication cost of $431.81/$\mathrm{kg_U}$, along with ten FTEs at $162,424.20/year, a fixed O&M cost of $0.676M/year, and a variable O&M cost of $2.12/MWh, the resulting LCOE is $55.25/MWh, which is not the lowest. This highlights that minimizing LCOE is a complex, interdependent optimization problem in which multiple cost factors interact nonlinearly. While OCC clearly plays a dominant role, achieving the lowest possible LCOE requires holistic optimization of both design and cost parameters tailored to each scenario. Changes in any one parameter can influence and alter the values of other interdependent parameters.

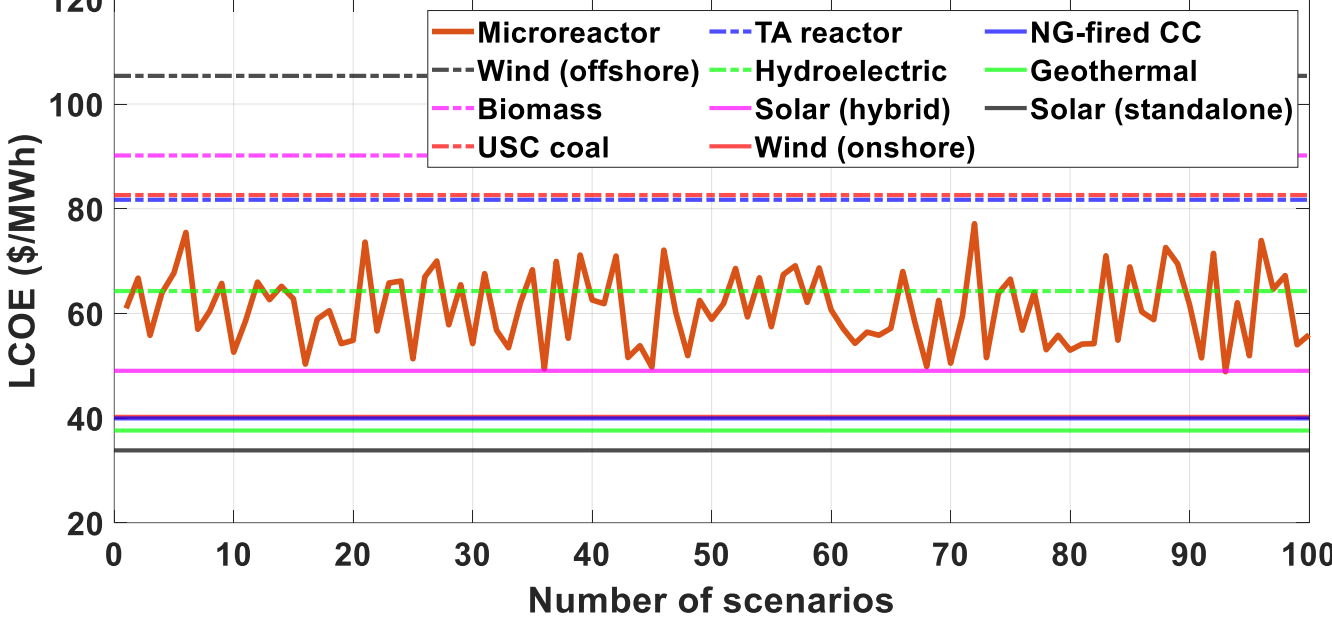

Fig. 4: Comparison of LCOE with different energy technologies for all scenarios considering all types of cost uncertainty

In this section, we explore the impact of uncertainty in OCC, O&M costs, and fuel costs separately. Table A2 (Appendix: A) and Figs A6-A10 (Appendix: B) presents the detailed statistical results when only OCC uncertainty is introduced into the optimization problem, while the other cost parameters are kept at their nominal values. The results show that LCOE can vary from $45.10/MWh to $66.15/MWh, with an SD of 5.51, presented in Figs. 5 and 8. This significant variation in LCOE is expected since OCC contributes the largest portion of LCOE, as shown in Fig. 2. However, the results indicate that microreactor LCOE remains cost-effective compared to offshore wind, biomass, USC coal, and TA reactors, as the maximum LCOE value obtained is $66.15/MWh, as shown in Fig. 5. The results also reveal that the PTC can reduce the LCOE by approximately 22.26% to 24.38%, even when accounting for a wide range of uncertainties in OCC.

Reactor capacity alters between 10.82 $\mathrm{MW_e}$ and 19.97 $\mathrm{MW_e}$ with a moderate SD value of 1.9, depicted in Fig. 9. Reactor capacity varies with changes in OCC, as the two parameters are interdependent. Fuel enrichment and discharge burnup values remain constant for all the simulation scenarios, as depicted in Figs. 10 and 13. Tail enrichment varies widely, similar to the previous case, presented in Fig. 11. Refueling duration varies from 9.88 years to 2.88 years, with a moderate SD of 1.88, shown in Fig. 12, similar to the case when all types of cost uncertainty are introduced.

The maximum LCOE of $66.15/MWh occurs at an OCC of $3939.39/kWe, while the minimum LCOE of $45.10/MWh corresponds to a lower OCC of $2515.15/kWe. As expected, a higher capital cost leads to a higher LCOE since all other cost parameters are held constant at their nominal values.

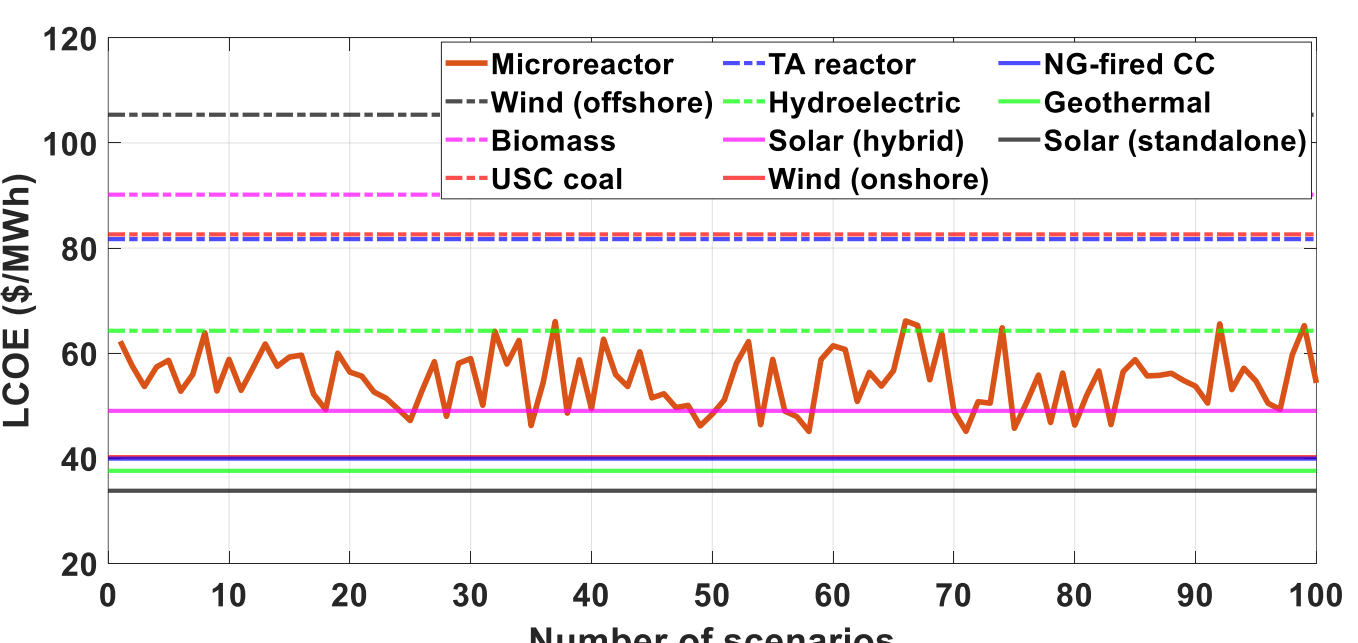

Fig. 5: Comparison of LCOE with different energy technologies for all scenarios considering OCC uncertainty

In this stage, we investigate the impact of O&M cost uncertainty on LCOE, while keeping the other cost parameters (e.g., OCC and fuel-related costs) at their nominal values. The O&M cost parameters include the number of FTEs for O&M, compensation per FTE, fixed O&M cost, and variable O&M cost. Table A3 (Appendix: A) and Figs. A11-A15 (Appendix: B) show the detailed optimization results with only O&M cost uncertainty introduced. The results are quite similar to those obtained when only OCC uncertainty was considered.

Although LCOE varies between $49.91/MWh and $66.94/MWh, with an SD of 3.05, microreactor LCOE becomes cost-effective compared to offshore wind, biomass, USC coal, and TA reactors regardless of the cost uncertainty, as depicted in Figs. 6 and 8. The minimum value of LCOE implies that microreactor can be cost-competitive than hybrid solar for some particular O&M cost parameters. Fuel enrichment and discharge burnup values remain unchanged due to the introduction of O&M cost uncertainty, as shown in Figs. 10 and 13. The results also reveal that the PTC can reduce the LCOE by approximately 22.66% to 24.38% when accounting for uncertainties in O&M costs.

The maximum LCOE of $66.94/MWh occurs when the number of FTEs for O&M is ten, with a compensation of $175,151.50/year, a fixed O&M cost of $0.531M/year, and a variable O&M cost of $2.44/MWh. In contrast, the minimum LCOE of $49.91/MWh is achieved with only three FTEs, compensated at $146,515.20/year, along with a fixed O&M cost of $0.471M/year, and a variable O&M cost of $2.06/MWh. These scenarios, with diverse levels of cost values, demonstrate that no single O&M cost component alone dictates the LCOE. Instead, it is the combined effect of all O&M-related parameters that determines cost efficiency. This underscores the fact that the optimal outcome results from the interplay among multiple variables rather than isolated changes in individual cost components.

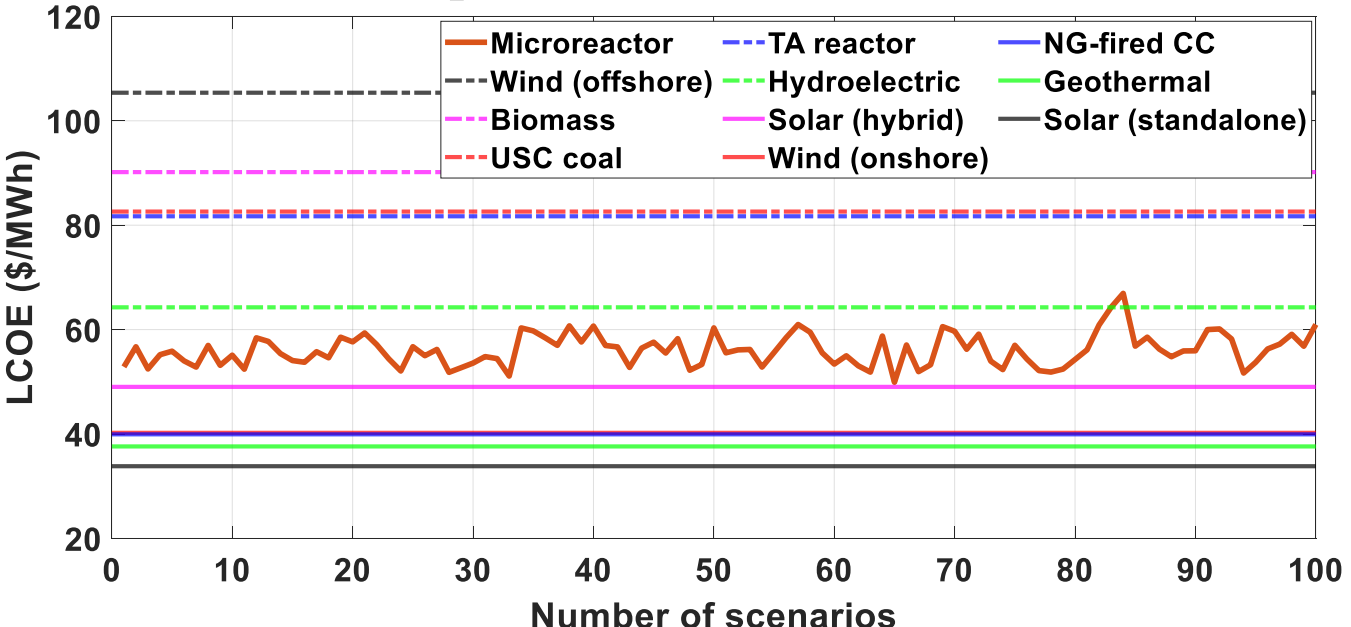

Fig. 6: Comparison of LCOE with different energy technologies for all scenarios considering O&M cost uncertainty

In this part, we assess the impact of fuel cost uncertainty on the LCOE. Fuel cost components include uranium cost, uranium conversion, uranium enrichment, and uranium fabrication cost. Table A4 (Appendix: A) and Figs. 7 and 8 show that LCOE varies between $49.72/MWh and $59.61/MWh. Thus, the LCOE remains more cost-effective than offshore wind, biomass, USC coal, TA reactors, and hydroelectric, shown in Fig.7. The slight variation in LCOE, depicted in Fig. 8, suggests that fuel cost has minimal impact, as indicated by the LCOE standard deviation and quartile values in Table A4. The results also indicate that PTC can reduce the LCOE by approximately 22.26% to 24.35% when considering uncertainties in fuel costs.

While fuel enrichment and discharge burnup show no variation, reactor capacity, tail enrichment, and refueling duration vary considerably, as shown in Table A4, Figs A16-A20, and Figs. 9-13. The overall results presented in Table A4 indicate that despite the uncertainty in fuel costs, the LCOE remains stable, highlighting the robustness of the microreactor's cost-effectiveness against fluctuations in fuel costs. The maximum LCOE of $59.61/MWh under fuel cost uncertainty occurs when the uranium cost is $93.45/lb, uranium conversion cost is $5.33/kg$_U$, uranium enrichment cost is $230.70/SWU, and fuel fabrication cost is $583.84/kg$_U$. On the other hand, the minimum LCOE of $49.72/MWh is observed when the uranium cost and uranium conversion cost are slightly higher at $96.36/lb and $6.97/kg$_U$, respectively, but the uranium enrichment and fabrication costs are significantly lower at $127.30/SWU and $467.17/kg$_U$, respectively. These results illustrate that individual fuel cost components, such as the market price of uranium, do not solely drive LCOE. Instead, it is the combined effect and interaction among all fuel cycle cost elements that shapes the final LCOE. For instance, lower enrichment or fabrication costs may offset a slightly higher uranium price, leading to a lower overall LCOE. This reinforces the view that minimizing LCOE requires a comprehensive optimization of the entire fuel cost structure, rather than focusing on any single parameter in isolation. While all types of cost uncertainty influence the LCOE (shown in Fig. 8), the results reveal a range of optimal design parameters within the explored design space, as illustrated in Figs. 9–13. These insights can serve as valuable guidelines for future microreactor designers.

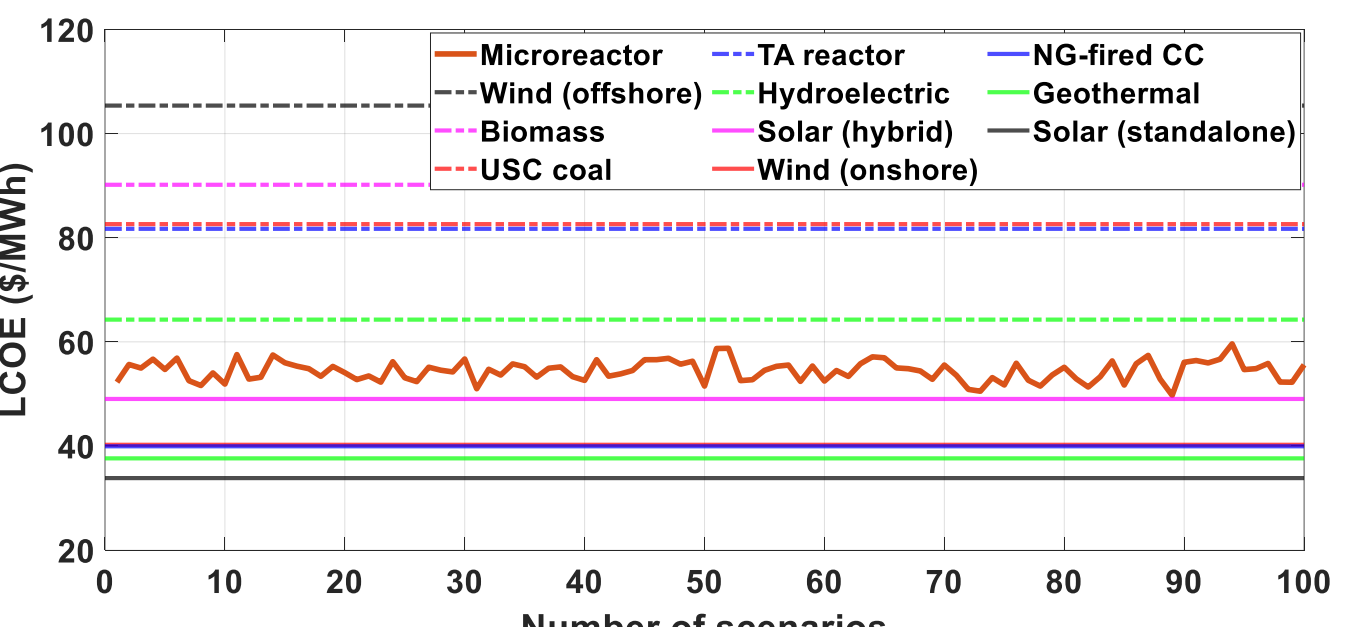

Fig. 7: Comparison of LCOE with different energy technologies for all scenarios considering fuel cost uncertainty

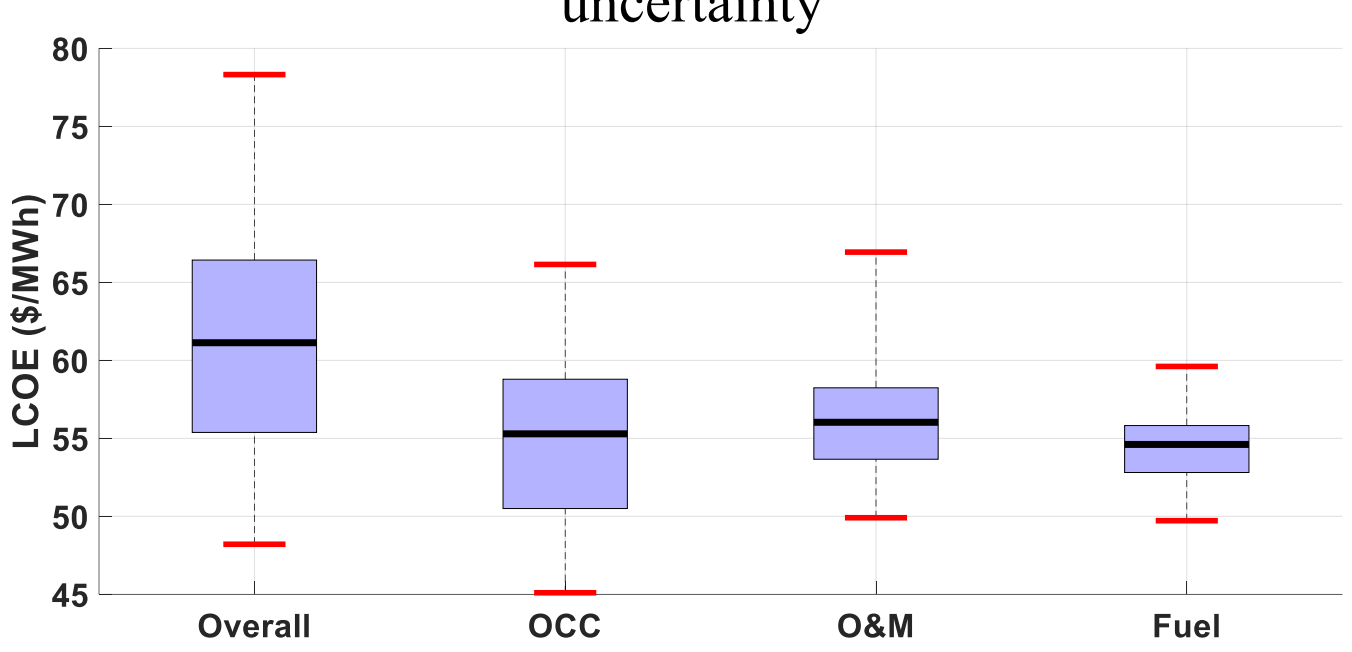

Fig. 8: Variation in LCOE while introducing all types of cost uncertainty (overall), OCC uncertainty, O&M cost uncertainty, fuel cost uncertainty.

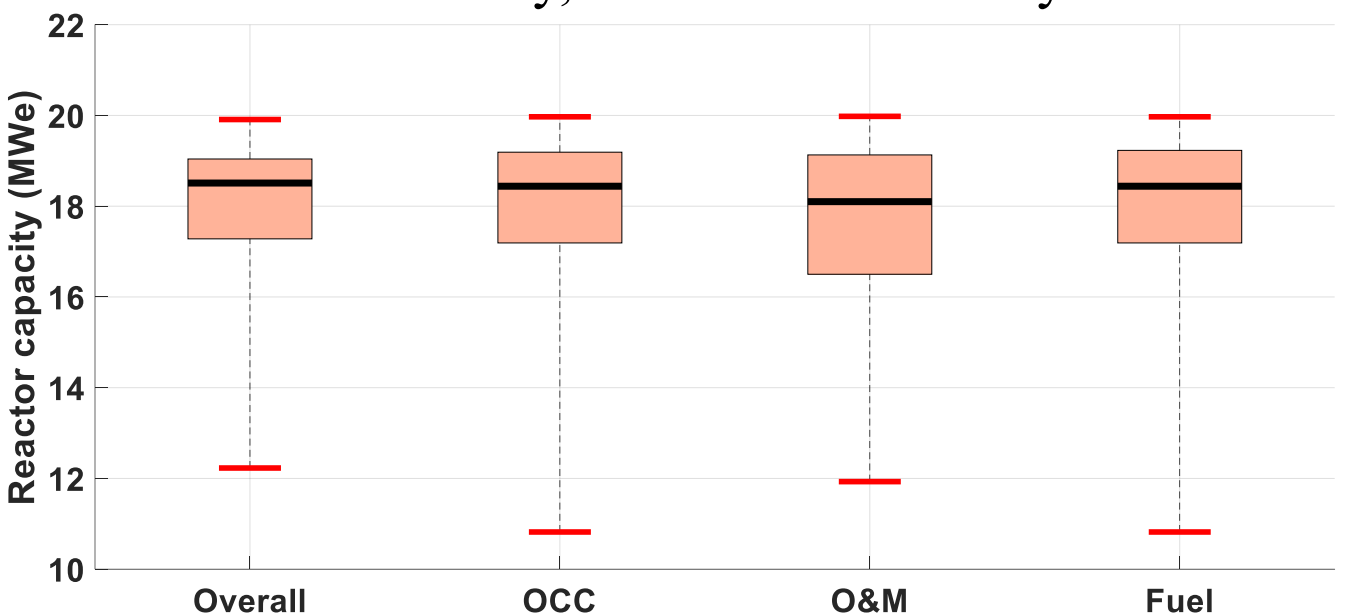

Fig. 9: Variation in reactor capacity while introducing all types of cost uncertainty (overall), OCC uncertainty, O&M cost uncertainty, fuel cost uncertainty.

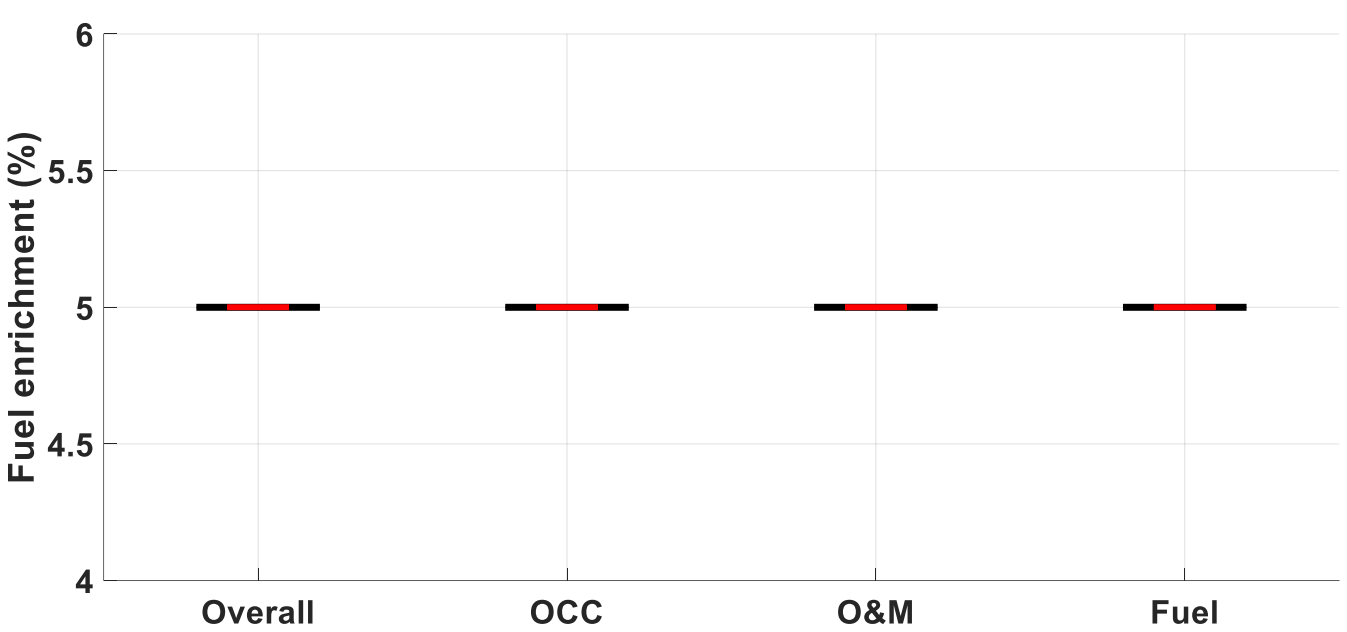

Fig. 10: Variation in fuel enrichment while introducing all types of cost uncertainty (overall), OCC uncertainty, O&M cost uncertainty, fuel cost uncertainty.

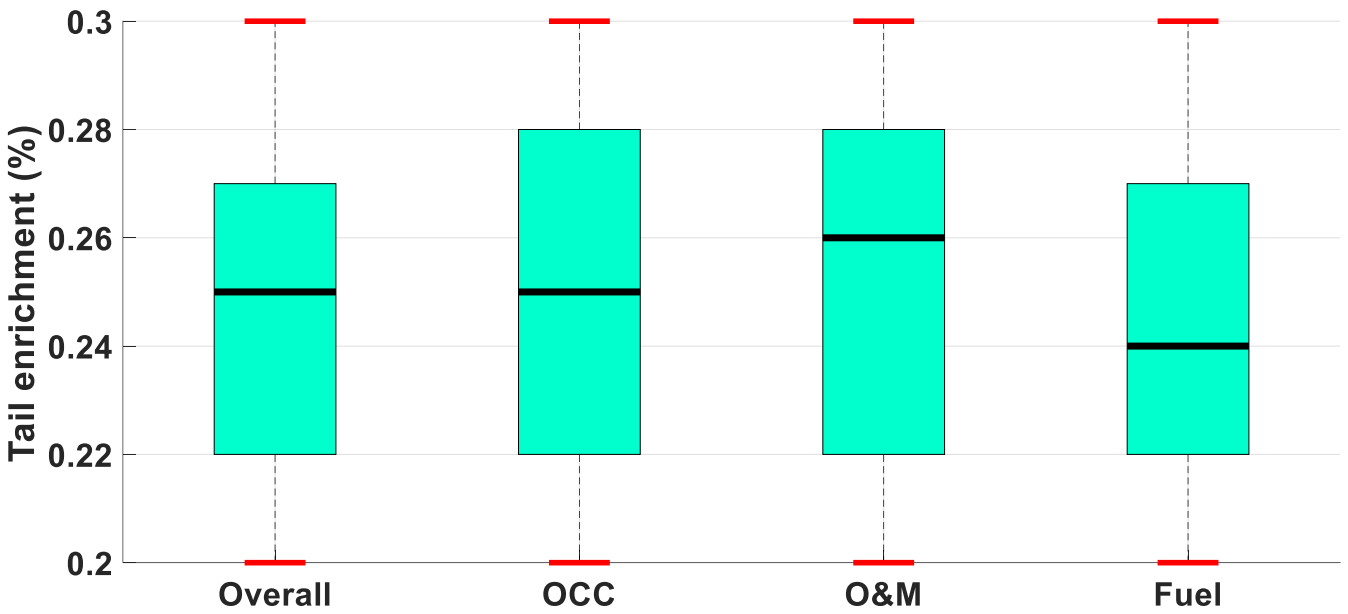

Fig. 11: Variation in tail enrichment while introducing all types of cost uncertainty (overall), OCC uncertainty, O&M cost uncertainty, fuel cost uncertainty.

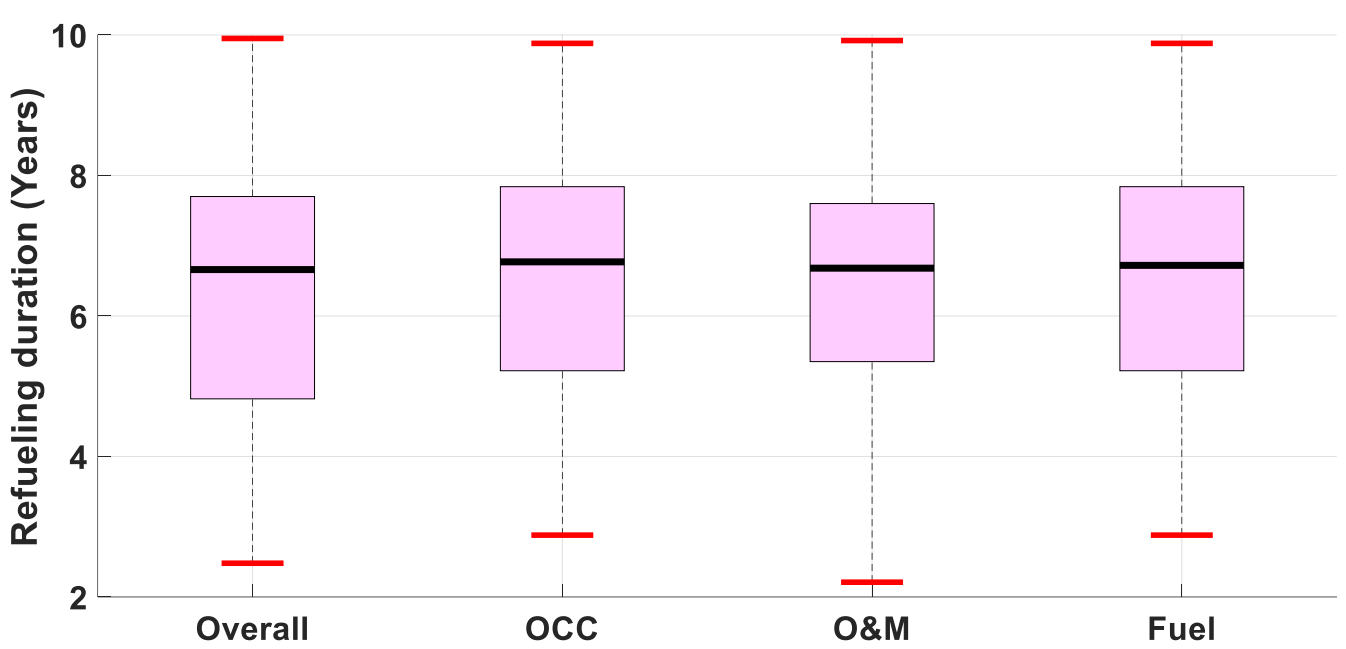

Fig. 12: Variation in refueling duration while introducing all types of cost uncertainty (overall), OCC uncertainty, O&M cost uncertainty, fuel cost uncertainty.

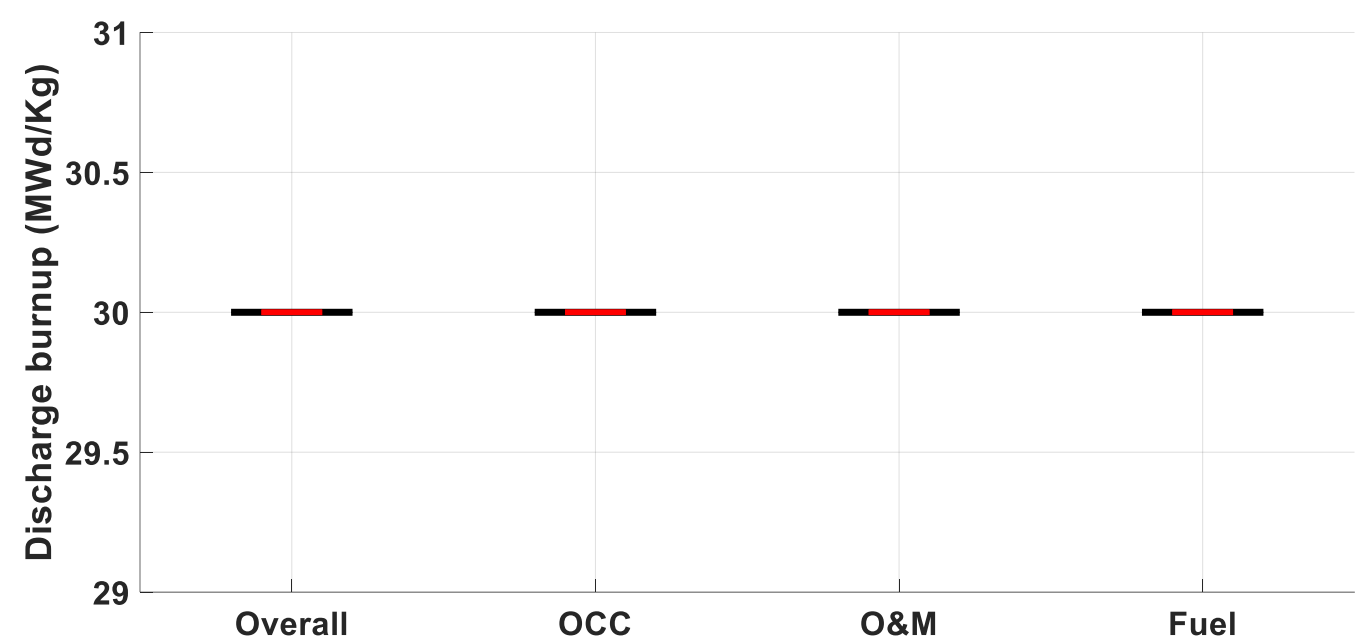

Fig. 13: Variation in discharge burnup while introducing all types of cost uncertainty (overall), OCC uncertainty, O&M cost uncertainty, fuel cost uncertainty.

Finally, we evaluate additional parameters, such as reactor thermal efficiency, real discount rate, and inflation rate, which are not assessed in previous sections. For the base case, reactor thermal efficiency is assumed to be 35%. However, advanced reactors, such as Fast Neutron Reactors (FNR), can achieve higher efficiencies. In this analysis, we vary the reactor thermal efficiency to 40%, 45%, and 50%. The results on Fig. 14 indicate that LCOE decreases with increased efficiency. This outcome is intuitive and can be mathematically explained using the equations in Section 3.1.

Figs. 15 and 16 illustrate the impact of the real discount rate and inflation rate on LCOE, respectively. The U.S. Department of Energy (DOE) sets the real discount rate and inflation rate for 2023 at 3% and 0.2%, respectively, for life-cycle cost analysis of capital investment projects for federal facilities, including energy efficiency, water conservation, sustainability, and resilience [34]. Hence, in this analysis, we assume lower values of 3% for the real discount rate and 0.2% for the inflation rate. The results show that LCOE decreases with lower real discount and inflation rates. For a 3% real discount rate and 0.2% inflation rate, the microreactor LCOE is $49.23/MWh and $49.04/MWh, respectively, which are comparable to hybrid solar. If both the real discount rate and inflation rate decline, the LCOE decreases further. For example, assuming a 3% real discount rate and a 0.2% inflation rate simultaneously, the microreactor LCOE would be $44.38/MWh. Therefore, the real discount rate and inflation rate are crucial factors in determining the LCOE of microreactors.

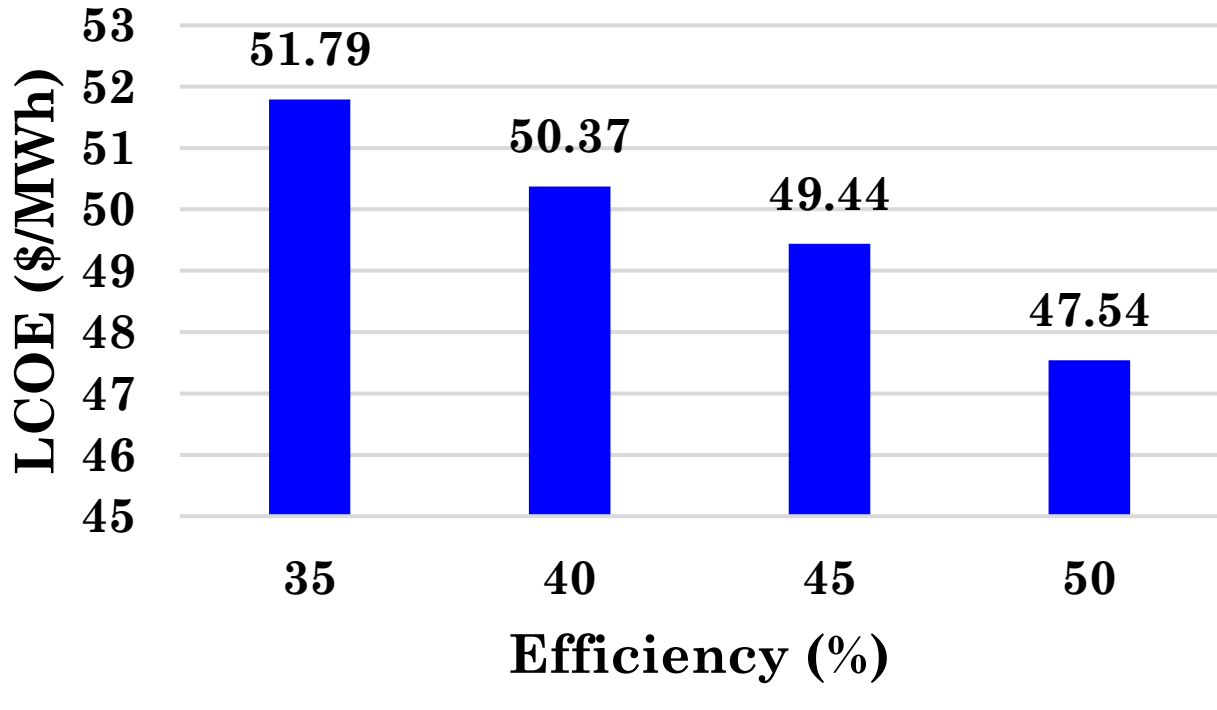

Fig. 14: Impact of reactor thermal efficiency

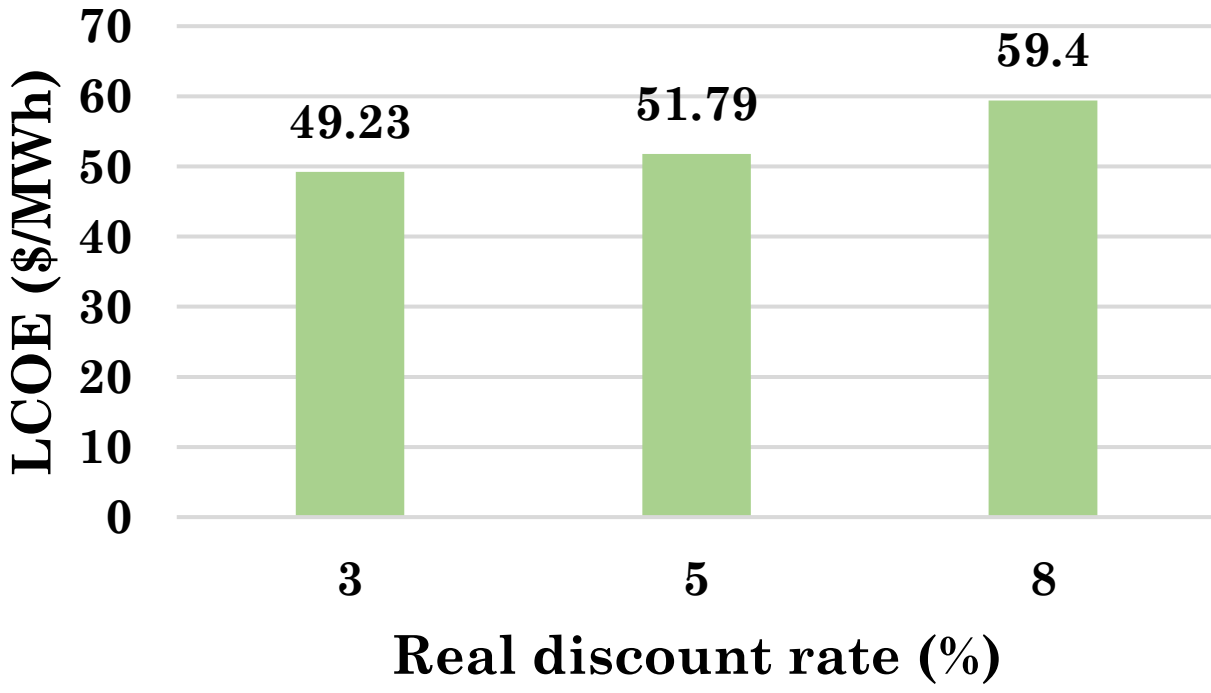

Fig. 15: Impact of real discount rate

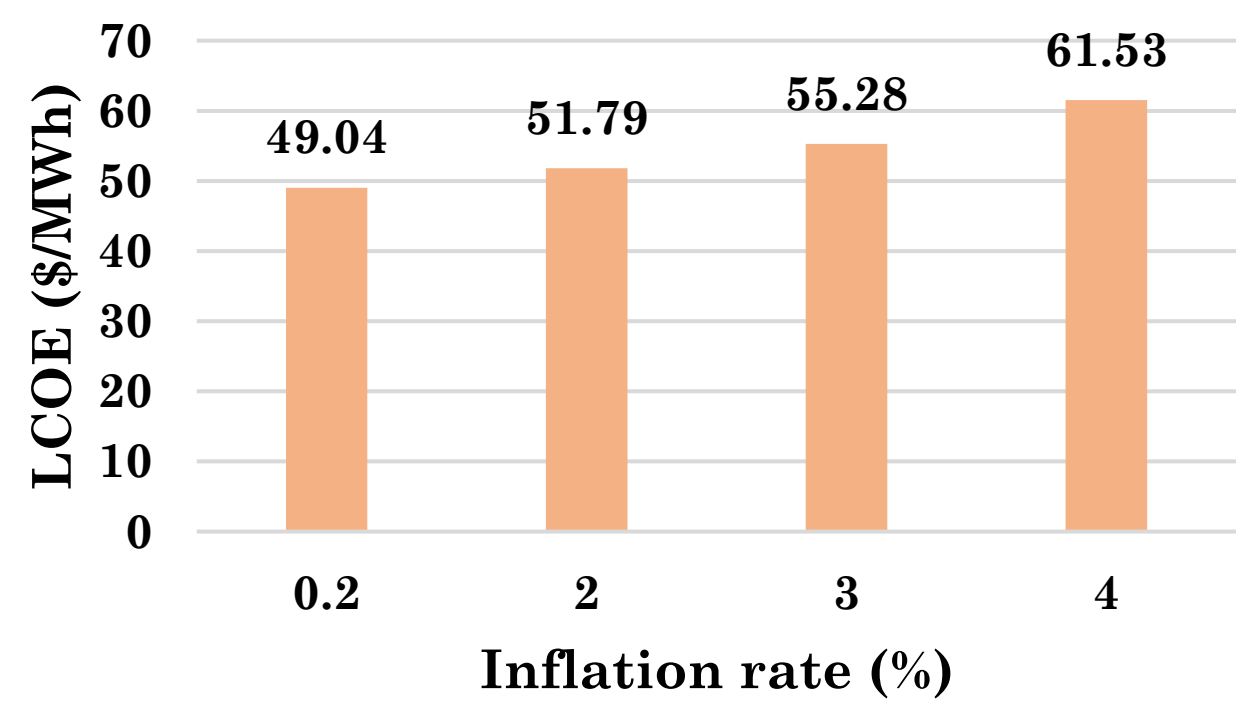

Fig. 16: Impact of inflation rate

## 5. CONCLUSION

This study provides a detailed analysis of microreactors' cost-competitiveness in the current electricity markets, focusing on the inherent uncertainties in reactor costs. By formulating 100 scenarios using PDFs of various reactor cost parameters and employing a GA coded in MATLAB, we evaluate the optimal technical parameters of the microreactor life cycle, including reactor capacity, fuel enrichment, tail enrichment, refueling time, and discharge burnup. While the use of 100 scenarios provides meaningful insight into cost variability, it is a limitation in terms of fully capturing the entire uncertainty space. The results demonstrate that microreactors can achieve competitive LCOE by optimizing these parameters, even considering cost uncertainties. Although there are no such microreactors having these parameters available in current markets, these results should serve as target benchmarks for microreactor designers aiming to achieve market cost competitiveness.

The base-case microreactor LCOE is calculated at $51.79/MWh when no cost uncertainty is considered in the optimization. When the impact of cost uncertainties is thoroughly assessed, it reveals that the LCOE for microreactors can range from $48.21/MWh to $78.32/MWh with a standard deviation of 7.2. The base case LCOE makes microreactors less competitive compared to hybrid solar, onshore wind, NG-fired CC, geothermal, and standalone solar while more suitable than offshore wind, biomass, USC coal, TA reactor, and hydroelectric. This variability underscores the importance of robust cost management and optimization strategies in microreactor deployment. The study also reveals the significant impact of the PTC on nuclear technologies. In particular, long-term policy certainty around tax incentives can play a critical role in de-risking investment and accelerating the deployment of advanced microreactors.

The study finds that the uncertainties in OCC and O&M costs have a more pronounced impact on LCOE than fuel costs. OCC contributes the most significant portion (58.3%) to the LCOE. Despite the high-cost sensitivity to OCC, microreactors remain cost-competitive with several existing energy technologies, including offshore wind, biomass, USC coal, and TA reactors, to some extent. On the other hand, O&M cost and fuel cost uncertainty has a less impact on LCOE. LCOE remains lower than offshore wind, biomass, USC coal, and TA reactor, regardless of O&M cost and fuel cost uncertainty. Contrary to common assumptions, the results show that O&M cost uncertainties have a more substantial impact on LCOE than fuel cost uncertainties. This counterintuitive finding challenges conventional reactor cost breakdowns. Additionally, optimal refueling duration, low fuel enrichment (5%), and high discharge burnup (30 MWd/kg) are crucial for enhancing cost efficiency. The quartile values from each statistical analysis provide further insights into the economic viability of microreactors despite cost uncertainties.

Future work could involve developing and utilizing more robust PDFs for the cost parameters. As microreactors become more prevalent, evaluating the uncertainty of more parameters impacting reactor economics will be easier. The findings of this study highlight the economic viability of microreactors as a competitive energy source. The optimization of fuel cycle parameters plays a pivotal role in achieving low LCOE, ensuring the economic competitiveness of microreactors despite inherent cost uncertainties. This study not only contributes valuable insights into the cost dynamics of microreactors but also sets a foundation for future research aimed at enhancing the economic efficiency of nuclear energy technologies. These results are instrumental in guiding policymakers and energy planners in making informed decisions regarding the integration of microreactors into the energy mix, fostering a sustainable and cost-effective energy future.

## ACKNOWLEDGEMENT

This work is sponsored by the Department of Energy Office of Nuclear Energy under project number (DE-NE0009382), which is funded through the Nuclear Energy University Program (NEUP).

## Appendix: A

Table A1: Optimal and statistical values of the objective function and decision variables considering all types of cost uncertainty

| **Parameters** | **Max** | **Min** | **SD** | **Q1** | **Median** | **Q3** |
|---|---|---|---|---|---|---|
| LCOE ($/MWh) | 78.32 | 48.21 | 7.2 | 55.38 | 61.13 | 66.43 |
| Reactor capacity $(MW_e)$ | 19.91 | 12.23 | 1.47 | 17.28 | 18.51 | 19.04 |
| Fuel enrichment (%) | 5 | 5 | 0 | 5 | 5 | 5 |
| Tail enrichment (%) | 0.3 | 0.2 | 0.03 | 0.22 | 0.25 | 0.27 |
| Refueling duration (Years) | 9.95 | 2.48 | 1.96 | 4.82 | 6.66 | 7.70 |
| Discharge burnup $(\text{MWd}/Kg)$ | 30 | 30 | 0 | 30 | 30 | 30 |

Table A2: Optimal and statistical values of the objective function and decision variables considering OCC uncertainty

| **Parameters** | **Max** | **Min** | **SD** | **Q1** | **Median** | **Q3** |
|---|---|---|---|---|---|---|
| LCOE ($/MWh) | 66.15 | 45.10 | 5.51 | 50.5 | 55.29 | 58.79 |
| Reactor capacity $(MW_e)$ | 19.97 | 10.82 | 1.9 | 17.19 | 18.44 | 19.19 |
| Fuel enrichment (%) | 5 | 5 | 0 | 5 | 5 | 5 |
| Tail enrichment (%) | 0.3 | 0.2 | 0.03 | 0.22 | 0.25 | 0.28 |
| Refueling duration (Years) | 9.88 | 2.88 | 1.88 | 5.22 | 6.77 | 7.84 |
| Discharge burnup $(\text{MWd}/Kg)$ | 30 | 30 | 0 | 30 | 30 | 30 |

Table A3: Optimal and statistical values of the objective function and decision variables considering O&M cost uncertainty

| **Parameters** | **Max** | **Min** | **SD** | **Q1** | **Median** | **Q3** |
|---|---|---|---|---|---|---|
| LCOE ($/MWh) | 66.94 | 49.91 | 3.05 | 53.66 | 56.03 | 58.24 |
| Reactor capacity $(MW_e)$ | 19.98 | 11.93 | 2.01 | 16.50 | 18.1 | 19.13 |
| Fuel enrichment (%) | 5 | 5 | 0 | 5 | 5 | 5 |
| Tail enrichment (%) | 0.3 | 0.2 | 0.03 | 0.22 | 0.26 | 0.28 |
| Refueling duration (Years) | 9.92 | 2.21 | 1.76 | 5.35 | 6.68 | 7.60 |
| Discharge burnup $(\text{MWd}/Kg)$ | 30 | 30 | 0 | 30 | 30 | 30 |

Table A4: Optimal and statistical values of the objective function and decision variables considering fuel cost uncertainty

| **Parameters** | **Max** | **Min** | **SD** | **Q1** | **Median** | **Q3** |
|---|---|---|---|---|---|---|
| LCOE ($/MWh) | 59.61 | 49.72 | 1.98 | 52.81 | 54.61 | 55.82 |
| Reactor capacity $(MW_e)$ | 19.97 | 10.82 | 1.93 | 17.19 | 18.44 | 19.23 |
| Fuel enrichment (%) | 5 | 5 | 0 | 5 | 5 | 5 |
| Tail enrichment (%) | 0.3 | 0.2 | 0.03 | 0.22 | 0.24 | 0.27 |
| Refueling duration (Years) | 9.88 | 2.88 | 1.88 | 5.22 | 6.72 | 7.84 |
| Discharge burnup $(\text{MWd}/Kg)$ | 30 | 30 | 0 | 30 | 30 | 30 |

## Appendix: B

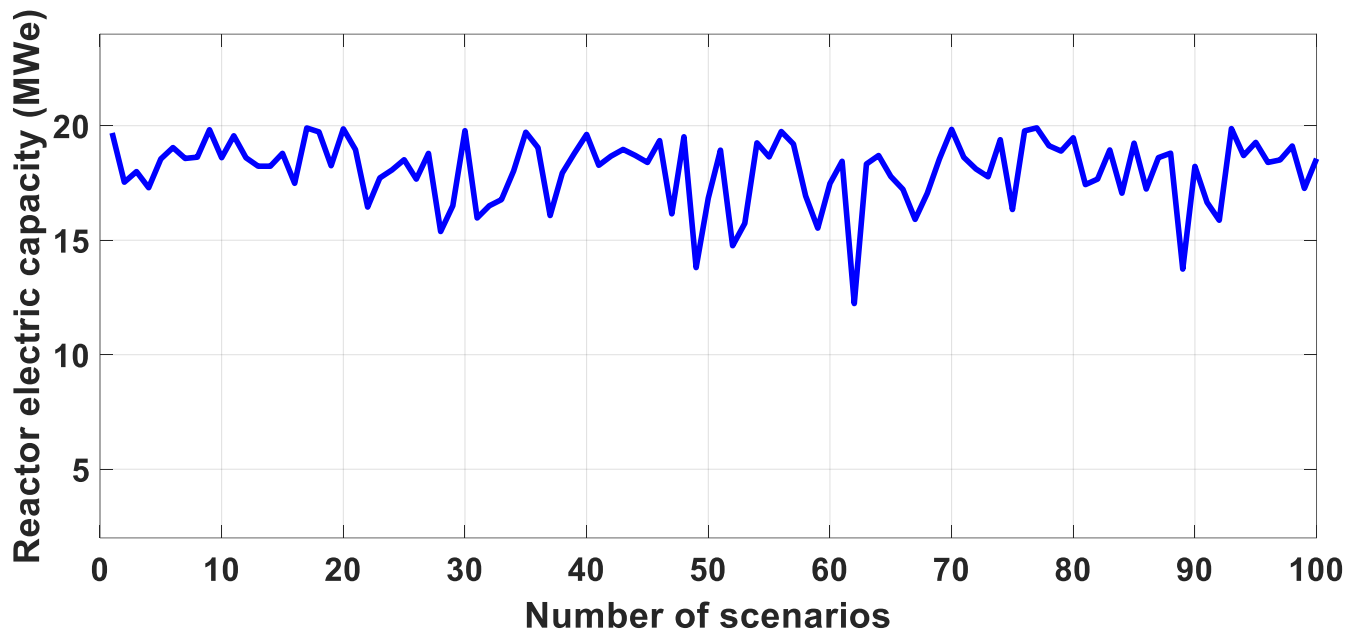

Fig A1: Optimal reactor capacity (electric) for all scenarios considering all types of cost uncertainty

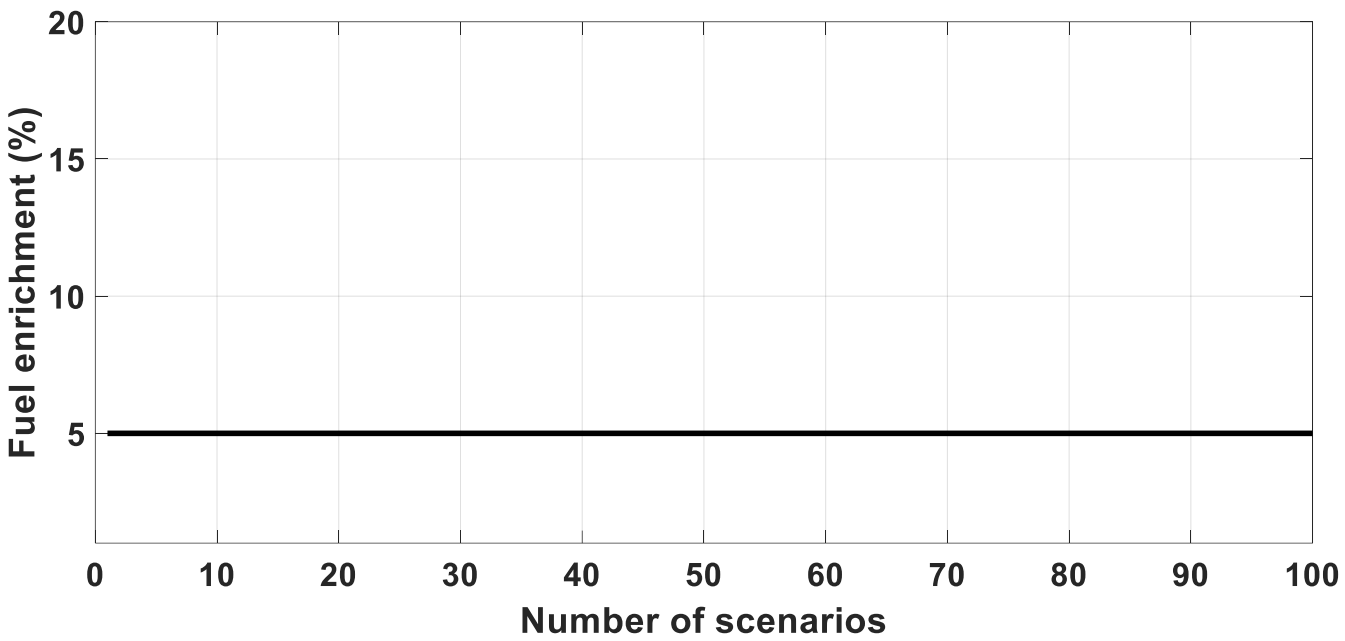

Fig A2: Optimal fuel enrichment for all scenarios considering all types of cost uncertainty

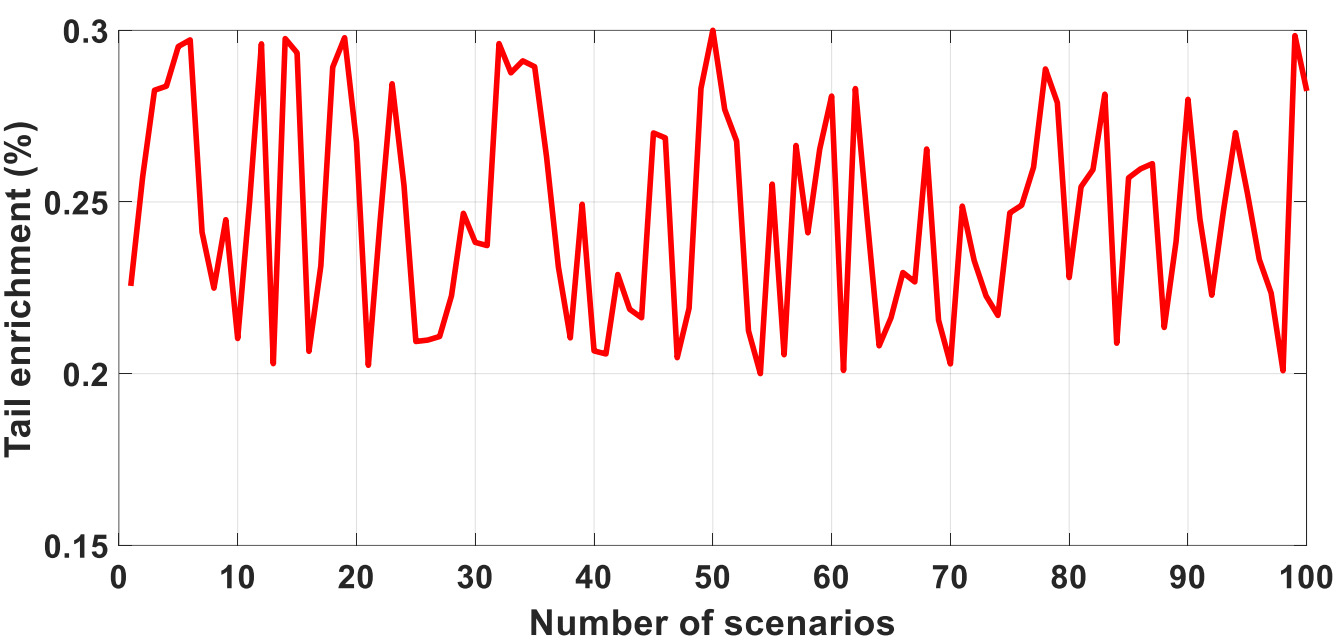

Fig A3: Optimal tail enrichment for all scenarios considering all types of cost uncertainty

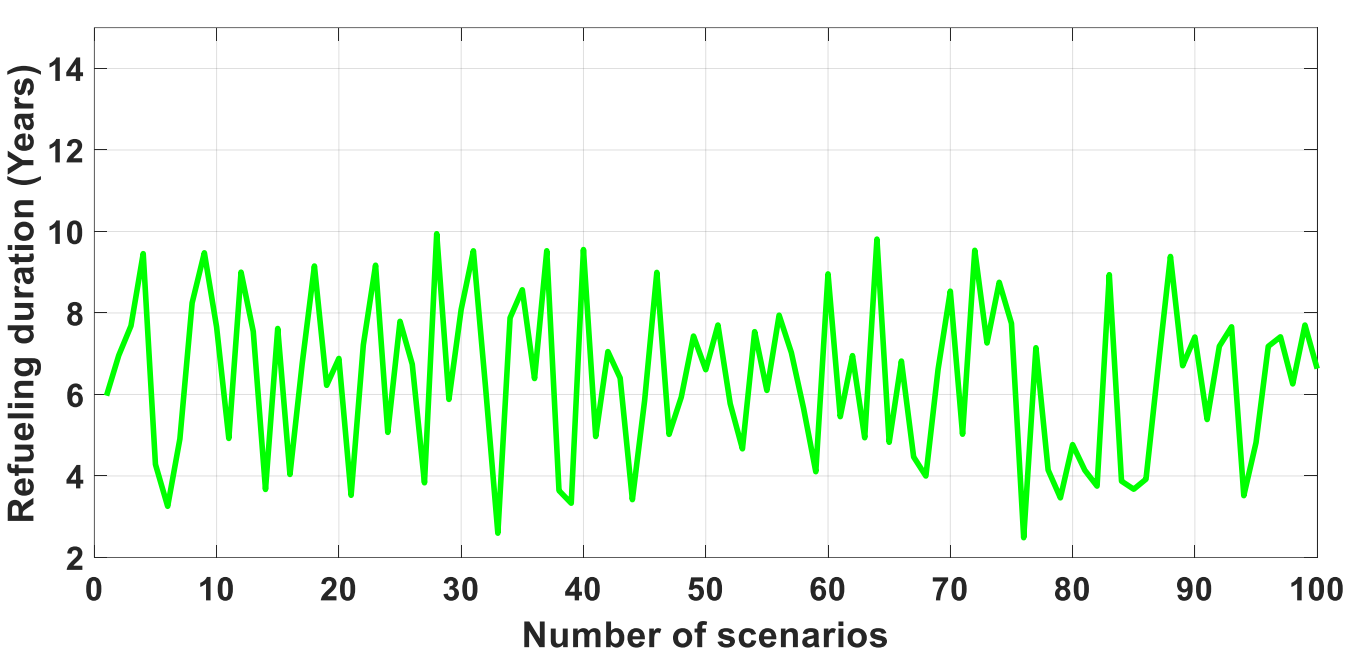

Fig A4: Optimal refueling duration for all scenarios considering all types of cost uncertainty

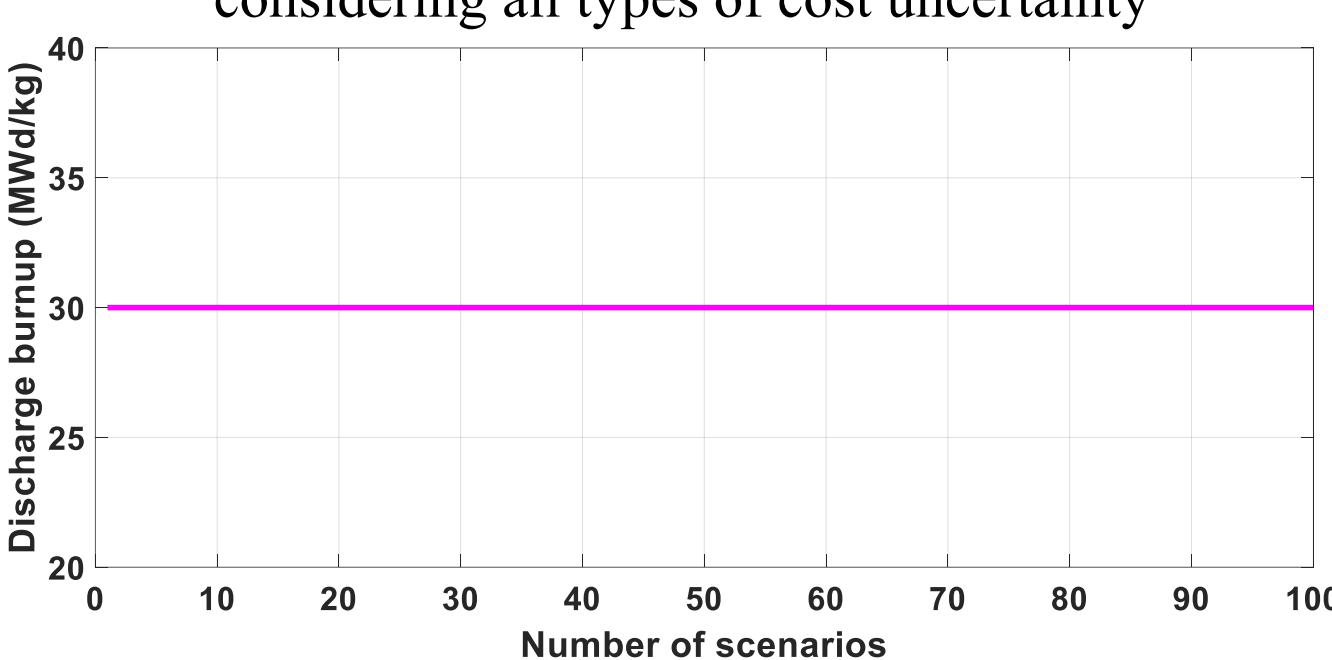

Fig A5: Optimal discharge burnup for all scenarios considering all types of cost uncertainty

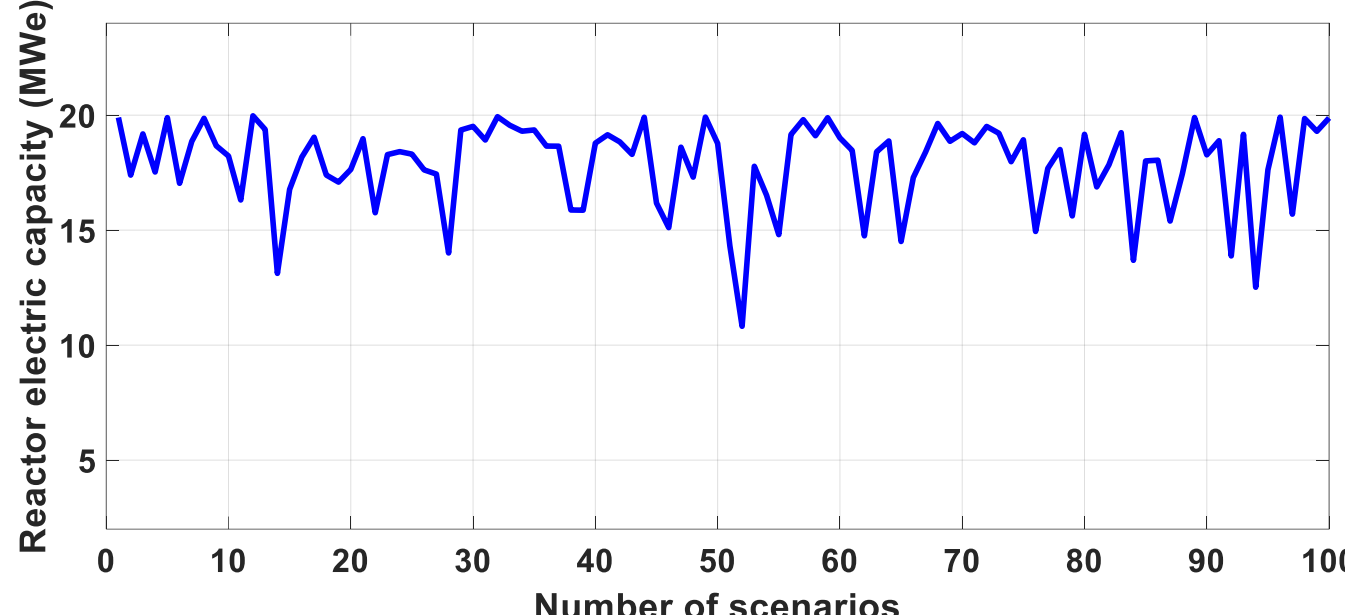

Fig A6: Optimal reactor capacity (electric) for all scenarios considering OCC uncertainty

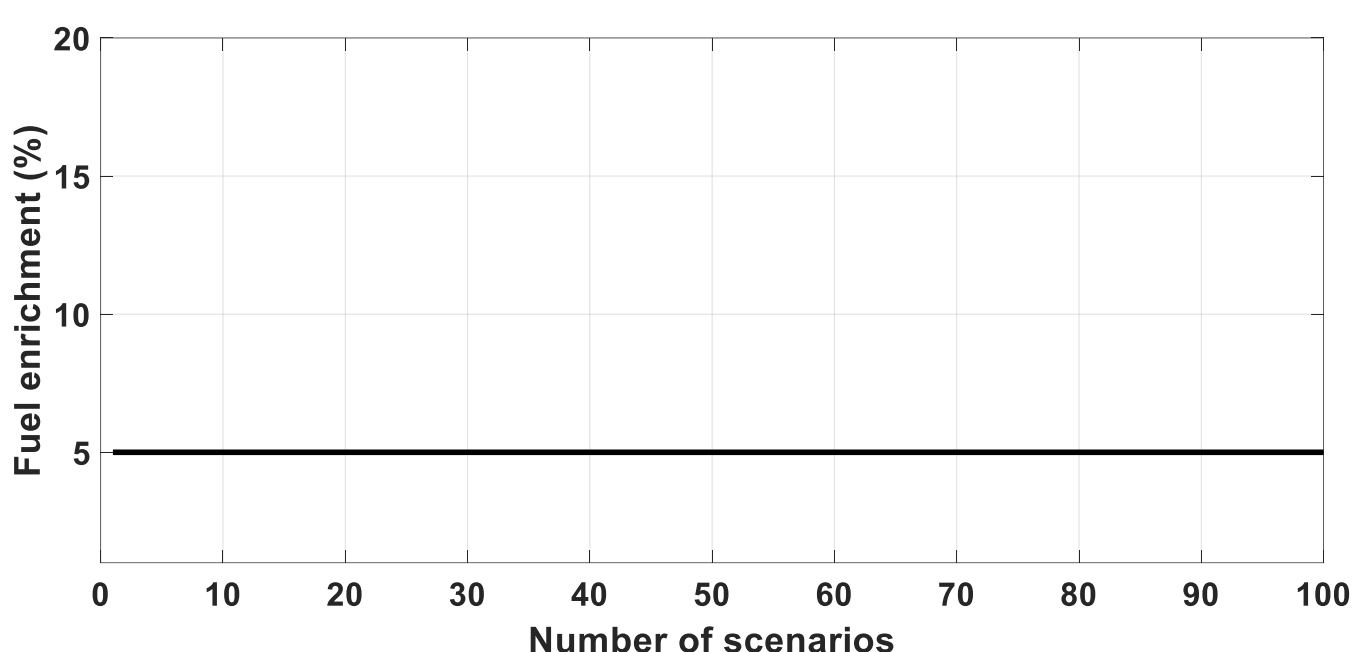

Fig A7: Optimal fuel enrichment for all scenarios considering OCC uncertainty

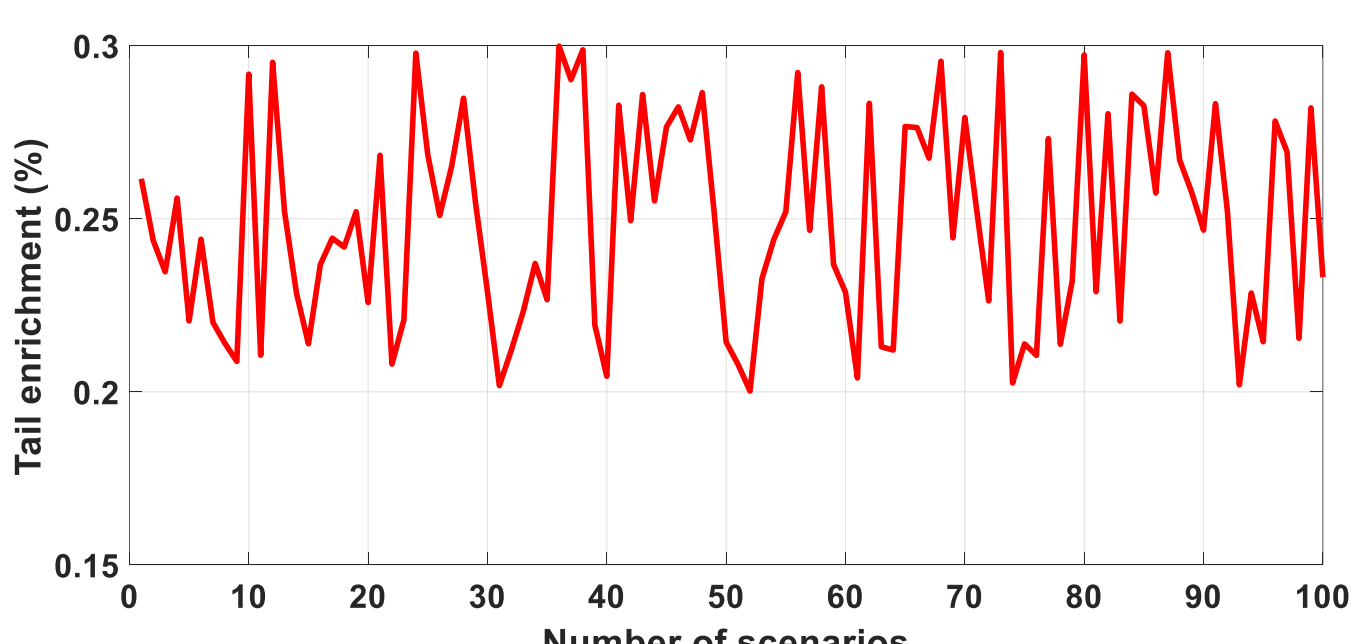

Fig A8: Optimal tail enrichment for all scenarios considering OCC uncertainty

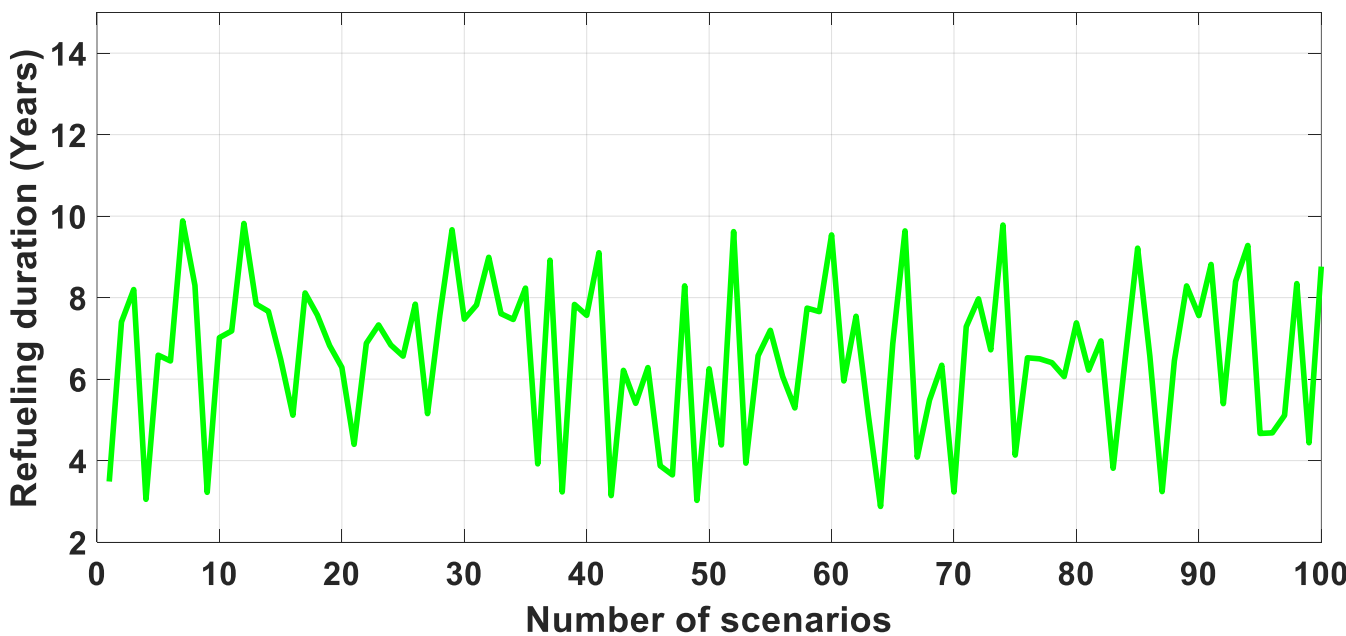

Fig A9: Optimal refueling duration for all scenarios considering OCC uncertainty

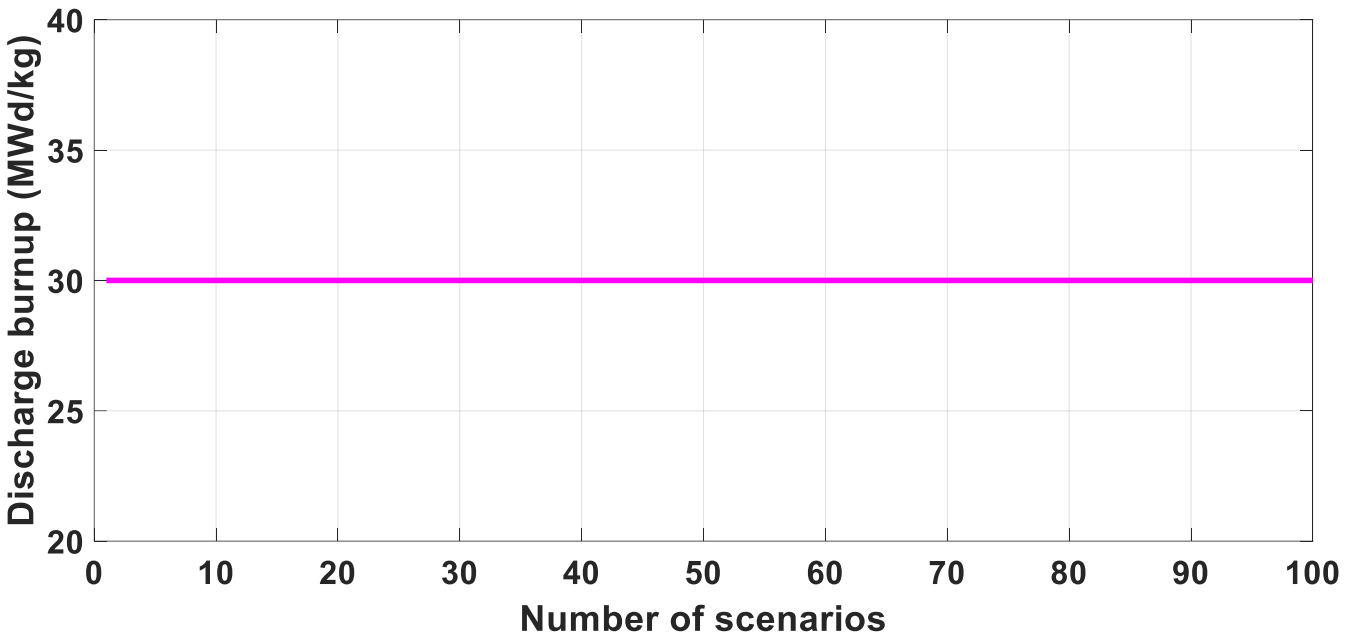

Fig A10: Optimal discharge burnup for all scenarios considering OCC uncertainty

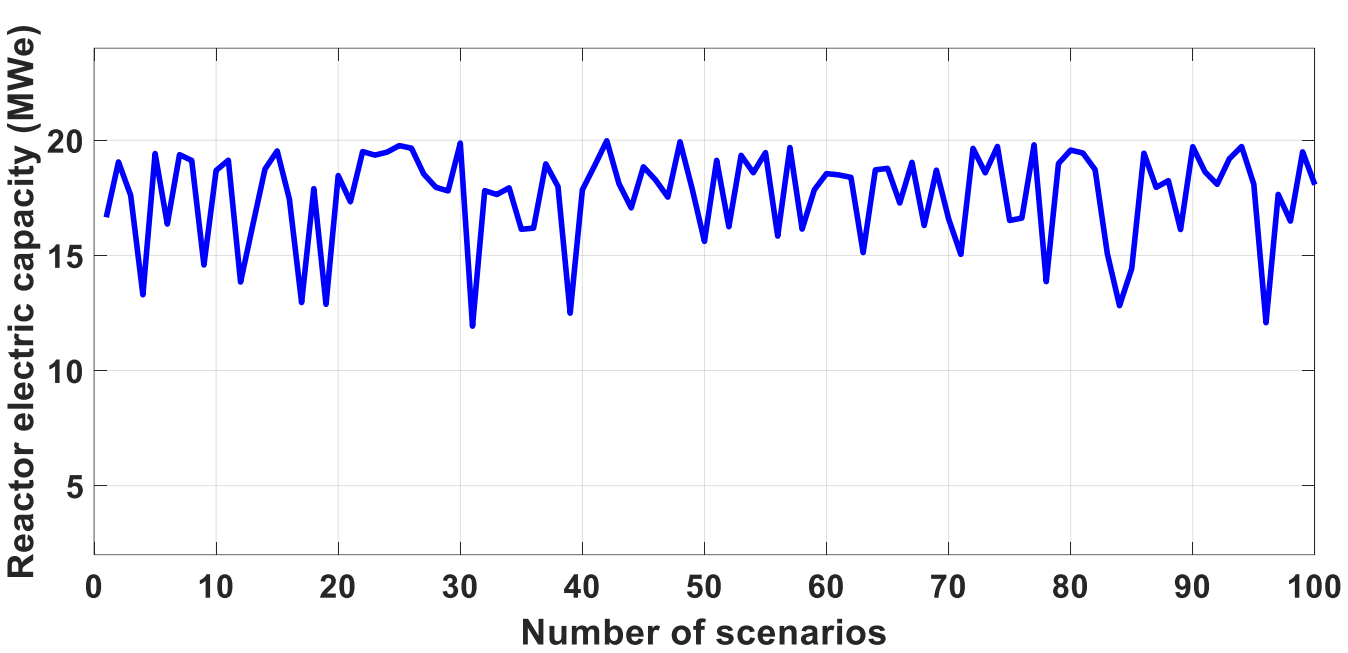

Fig A11: Optimal reactor capacity (electric) for all scenarios considering O&M cost uncertainty

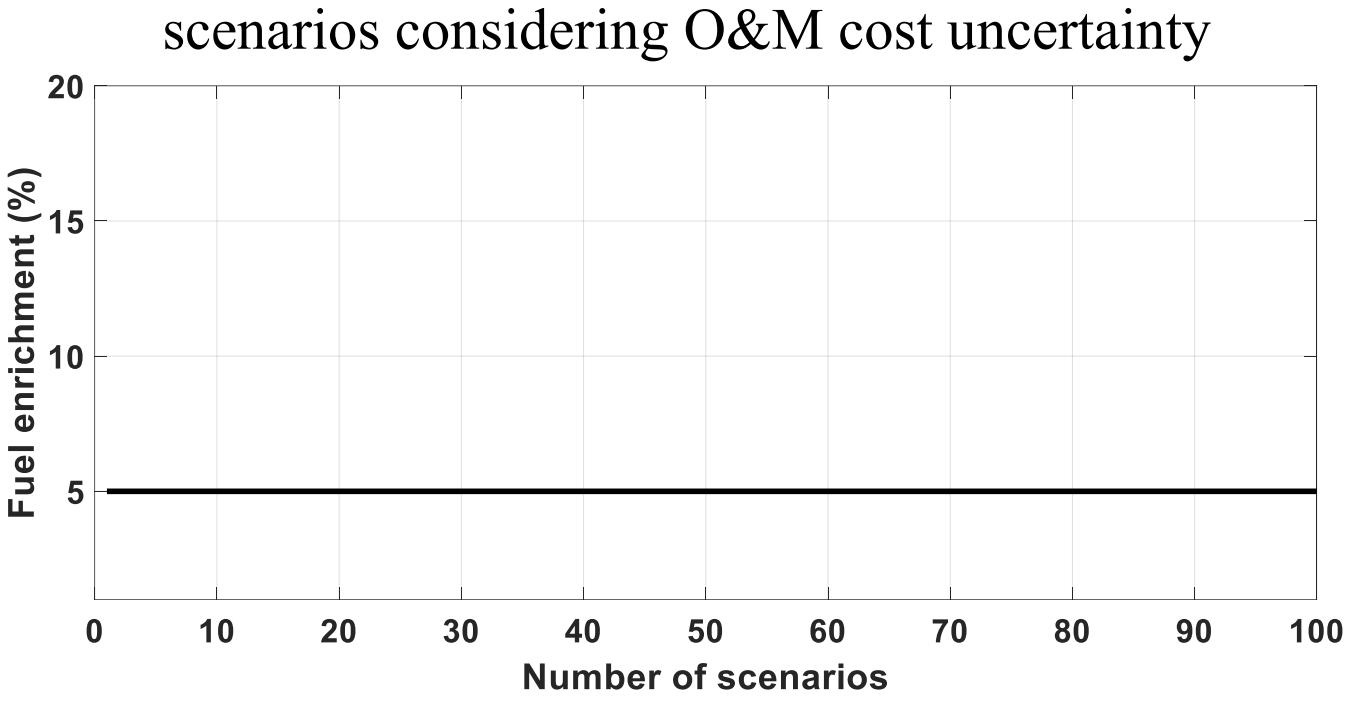

Fig A12: Optimal fuel enrichment for all scenarios considering O&M cost uncertainty

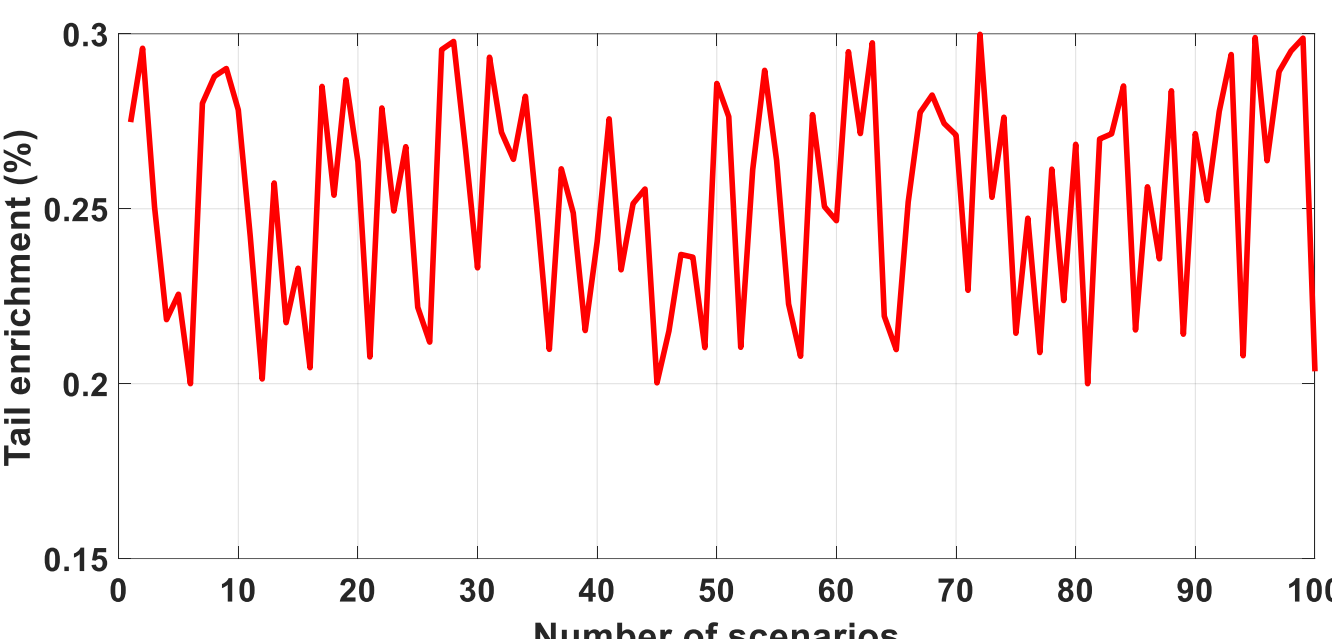

Fig A13: Optimal tail enrichment for all scenarios considering O&M cost uncertainty

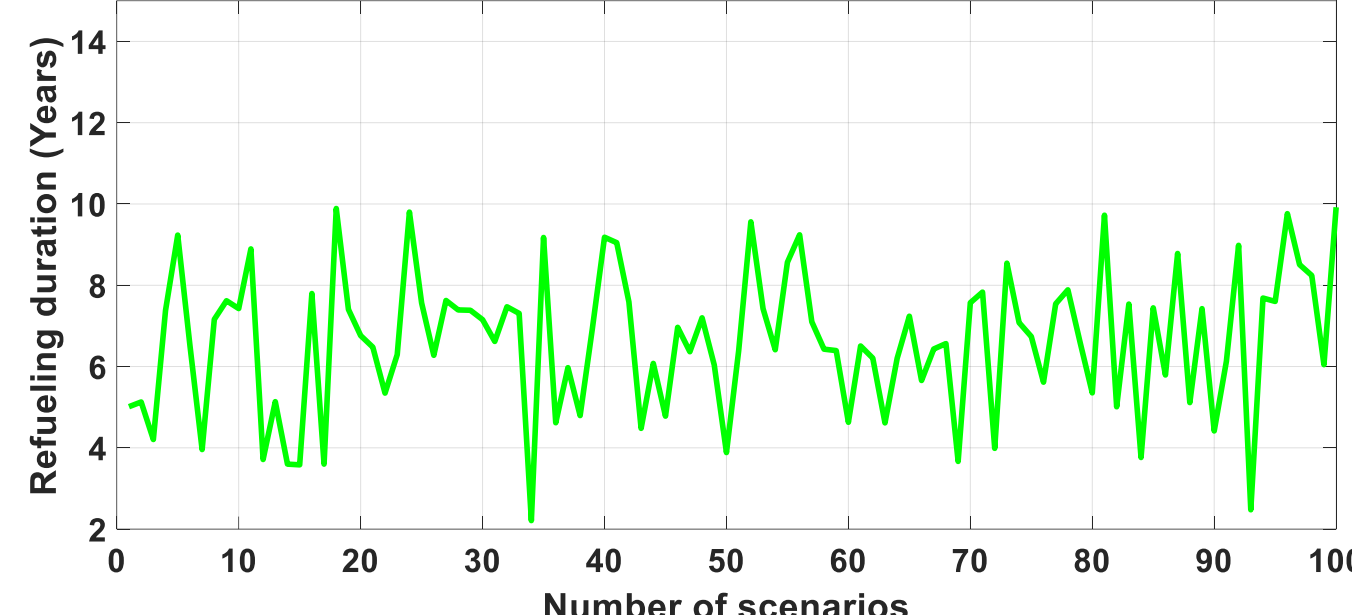

Fig A14: Optimal refueling duration for all scenarios considering O&M cost uncertainty

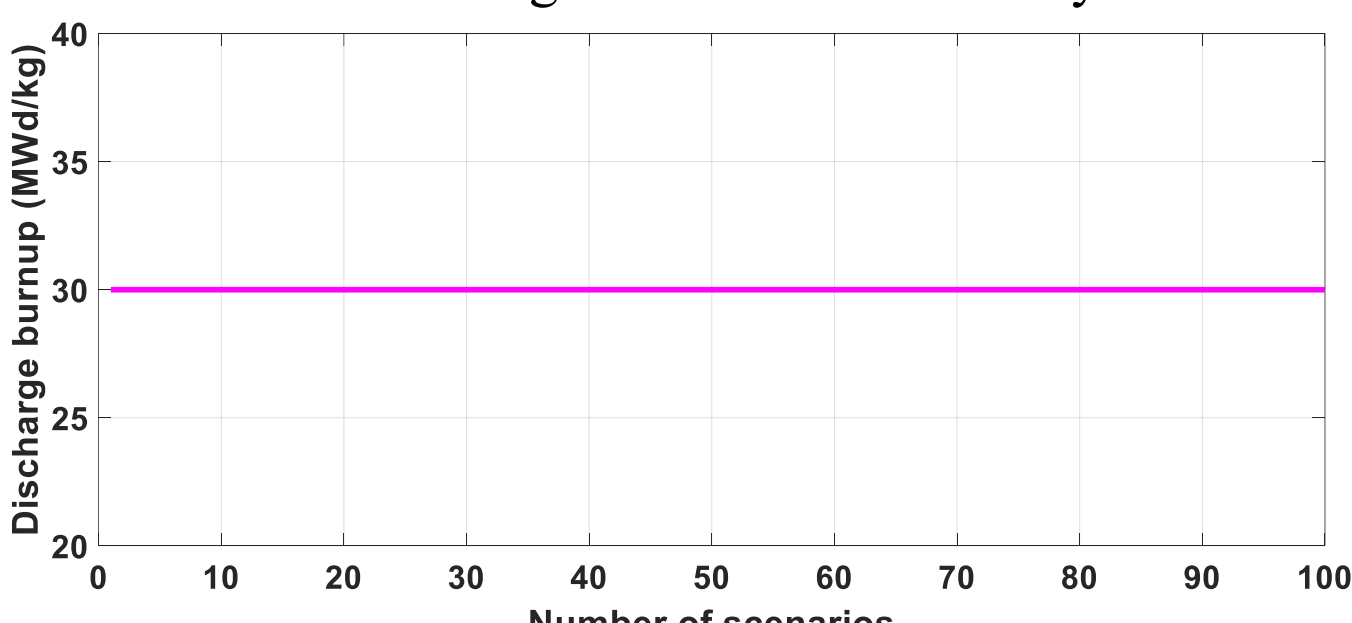

Fig A15: Optimal discharge burnup for all scenarios considering O&M cost uncertainty

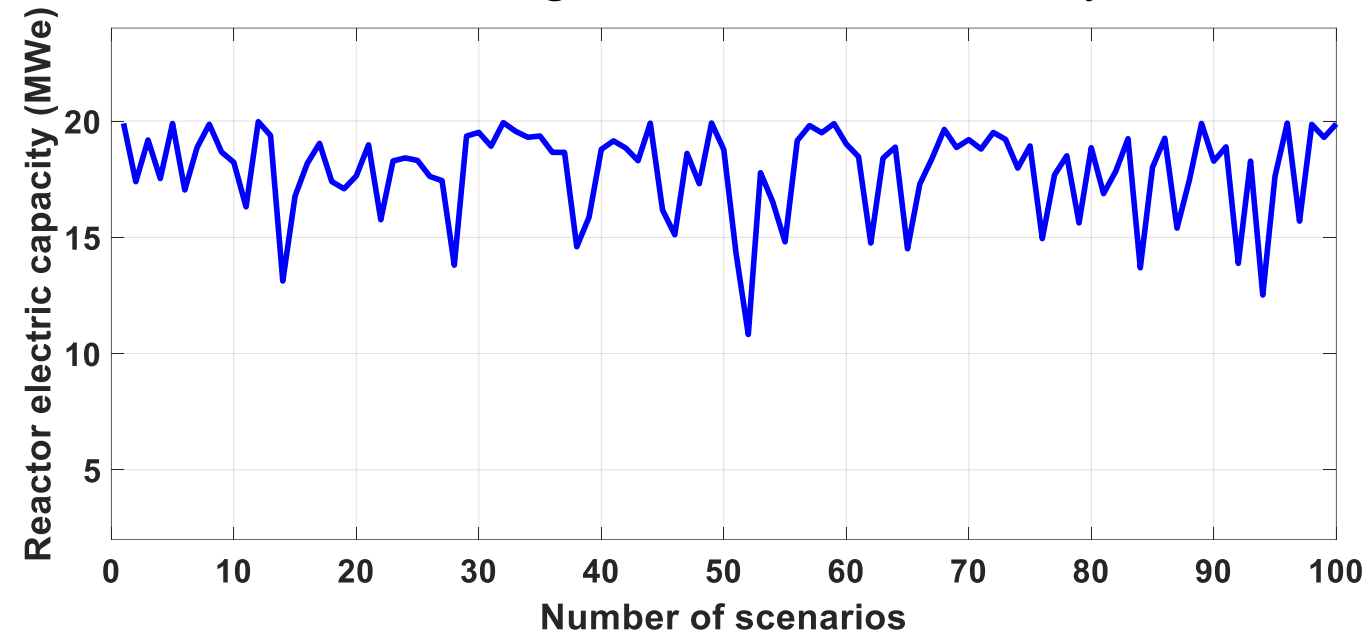

Fig A16: Optimal reactor capacity (electric) for all scenarios considering fuel cost uncertainty

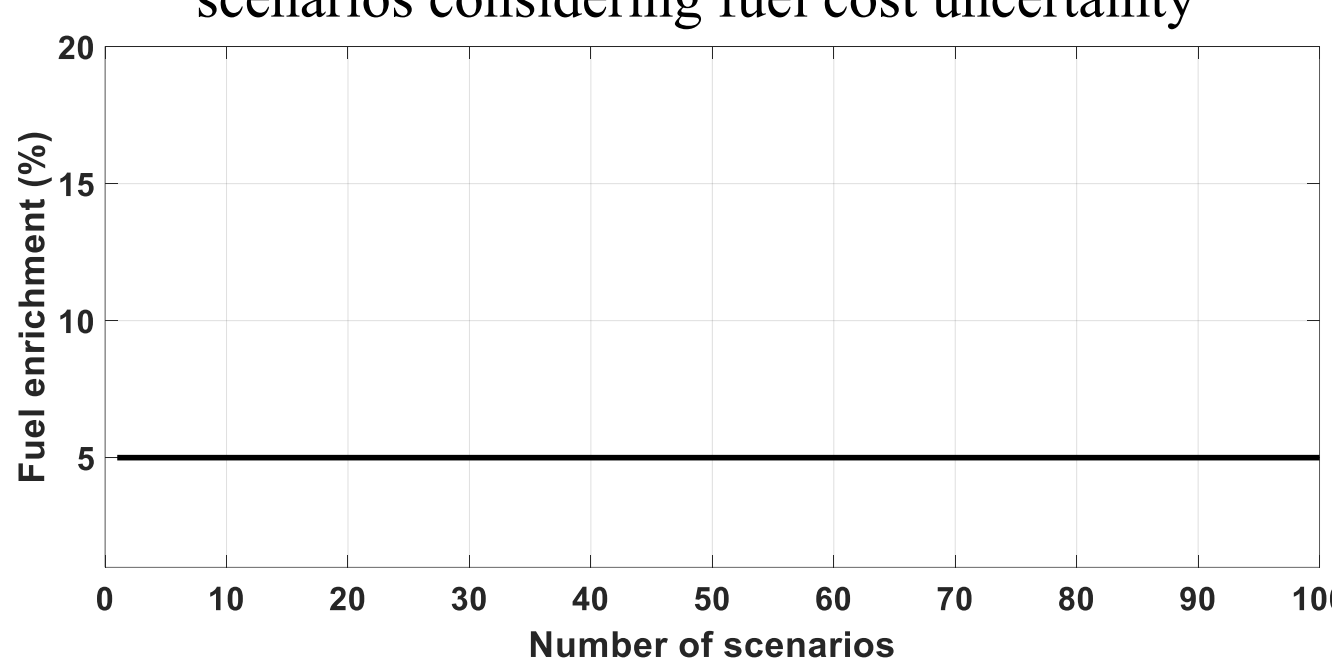

Fig A17: Optimal fuel enrichment for all scenarios considering fuel cost uncertainty

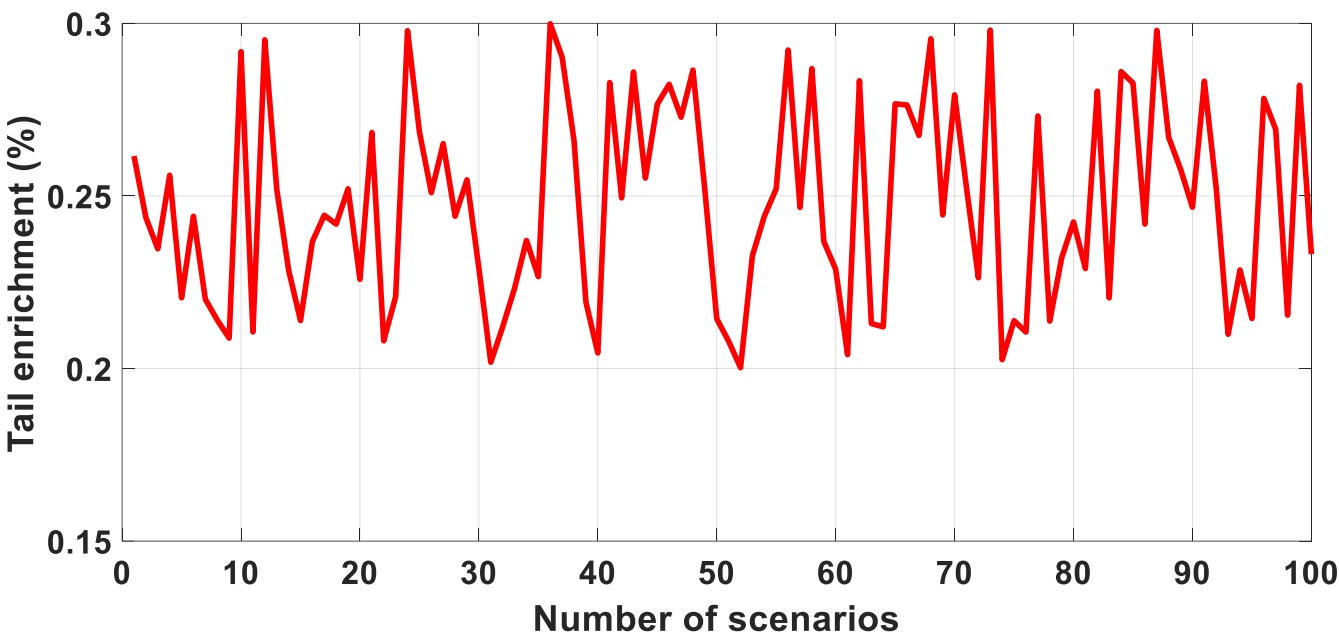

Fig A18: Optimal tail enrichment for all scenarios considering fuel cost uncertainty

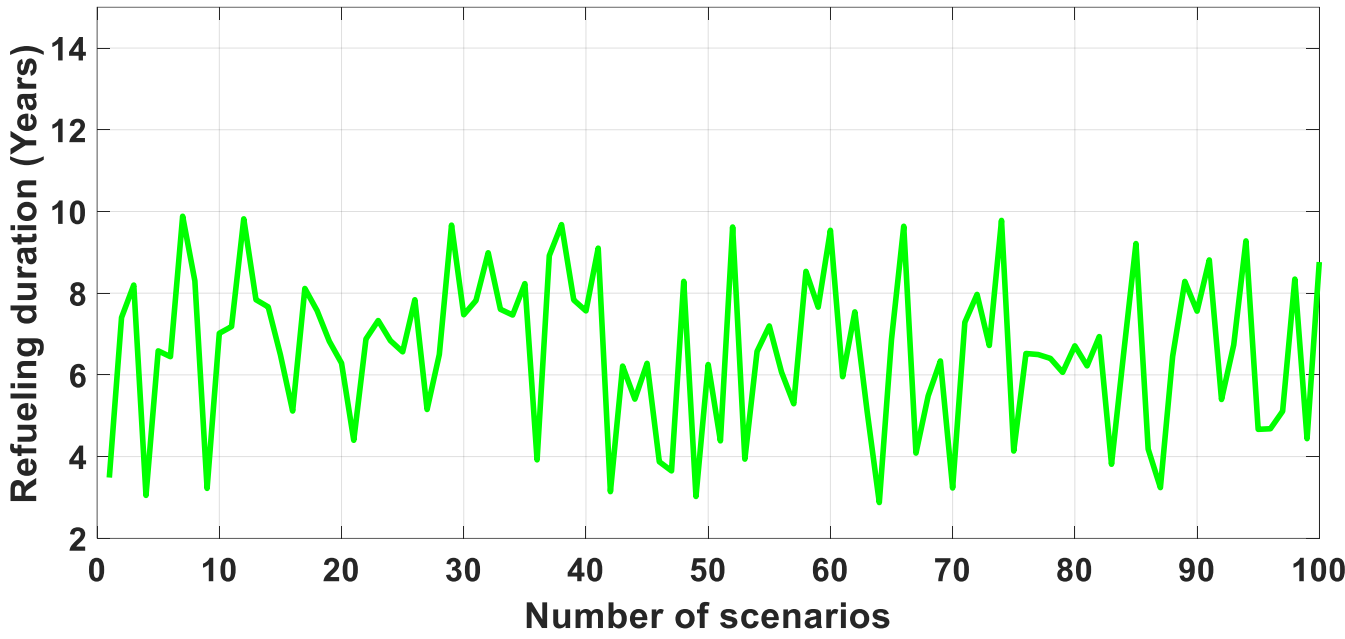

Fig A19: Optimal refueling duration for all scenarios considering fuel cost uncertainty

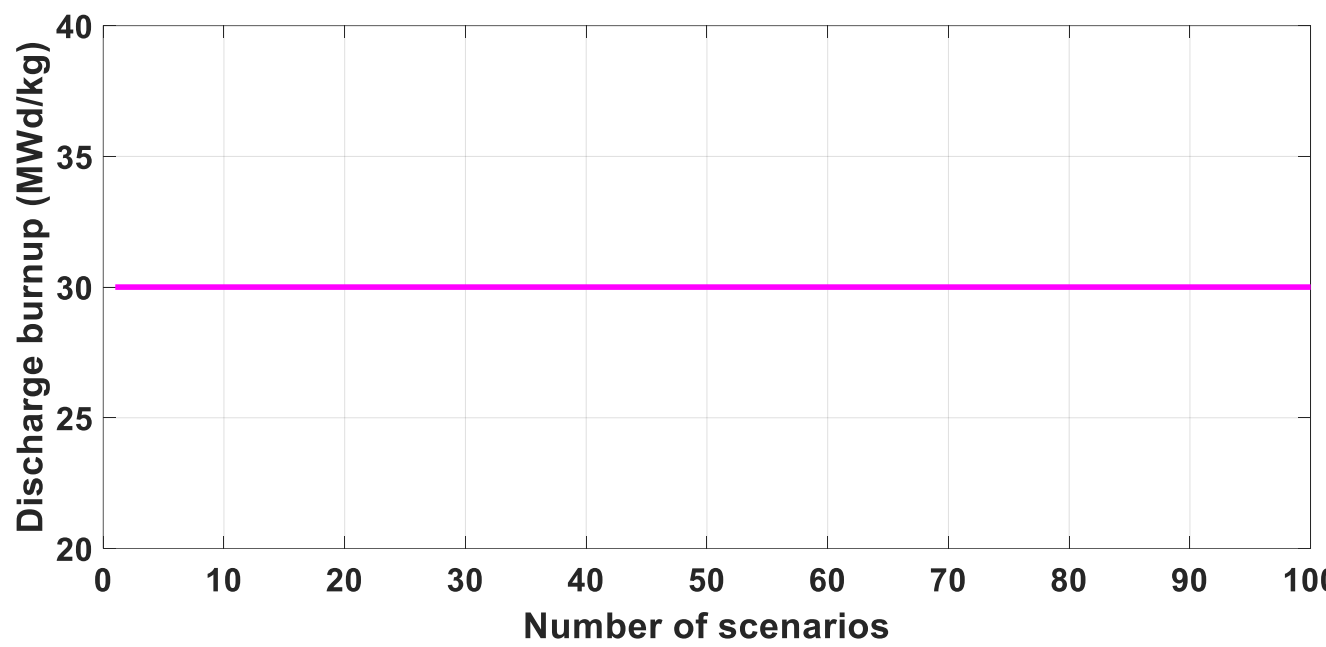

Fig A20: Optimal discharge burnup for all scenarios considering fuel cost uncertainty